\documentclass[10pt,twocolumn,letterpaper]{article}
\pdfoutput=1

\usepackage{cvpr}
\usepackage{times}
\usepackage{epsfig}
\usepackage{graphicx}
\usepackage{amsmath}
\usepackage{amssymb}
\usepackage{color}
\usepackage{url}

\usepackage{multirow}

\usepackage{booktabs}

\usepackage{pifont}

\usepackage[pagebackref=true,breaklinks=true,letterpaper=true,colorlinks,bookmarks=false]{hyperref}

\newcommand{\BestVOCSevenTest}{79.2}
\newcommand{\BestVOCTwelveTest}{76.4}
\newcommand{\BestCOCOTest}{24.9}

\definecolor{kb_color}{rgb}{1,.14,0}
\definecolor{larry_color}{rgb}{0.2,.64,0}
\definecolor{ross_color}{rgb}{.6,.25,.5}
\definecolor{sean_color}{rgb}{1,.5,0}
\definecolor{todo_color}{rgb}{.8,.8,.1}
\definecolor{new_color}{rgb}{.9,.1,.1}

\newcommand{\newtext}[1]{#1}

\newenvironment{packed_enum}{
\begin{enumerate}
  \setlength{\itemsep}{1pt}
  \setlength{\parskip}{0pt}
  \setlength{\parsep}{0pt}
}{\end{enumerate}}

\newcommand{\leaveout}[1]{}

\usepackage{amsopn}

\def\cvprPaperID{1646} % *** Enter the CVPR Paper ID here

\cvprfinalcopy
\begin{document}

%\title{Detecting Objects in Context with Multi-layer ROI Pooling and Recurrent
%Neural Networks.}
%\title{Detecting Objects in Context with the Inside-Outside Net}
\title{%
  \vspace{-0.43em}
  Inside-Outside Net:
  Detecting Objects in Context with Skip Pooling and
  Recurrent Neural Networks
  \vspace{-0.43em}
}
%Skip Pooling and Recurrent Neural Networks}

\author{%
Sean Bell$^1$
\qquad
C. Lawrence Zitnick$^2$
\qquad
Kavita Bala$^1$
\qquad
Ross Girshick$^2$
\\
$^1$Cornell University
\qquad
$^2$Microsoft Research\thanks{\newtext{Ross Girshick and C. Lawrence Zitnick are now at Facebook AI Research.}}\\
{\tt\small \{sbell,kb\}@cs.cornell.edu}
\qquad
{\tt\small rbg@fb.com}
}

\maketitle

\begin{abstract}
It is well known that contextual and multi-scale representations are important
for accurate visual recognition.  In this paper we present the Inside-Outside
Net (ION), an object detector that exploits information both inside and outside
the region of interest. Contextual information outside the region of interest is
integrated using spatial recurrent neural networks. Inside, we use skip pooling
to extract information at multiple scales and levels of abstraction.  Through
extensive experiments we evaluate the design space and provide readers with an
overview of what tricks of the trade are important.  ION improves
state-of-the-art on PASCAL VOC 2012 object detection from 73.9\% to 76.4\% mAP.
On the new and more challenging MS COCO dataset, we improve state-of-art-the
from 19.7\% to \newtext{33.1\% mAP}.  \newtext{In the 2015 MS COCO Detection Challenge,
our ION model won the Best Student Entry and finished $3^\text{rd}$ place overall.}
  %We further show an improvement to 32.3\%
%with improved box proposals, left-right flips, weighted voting, and optimized
%hyperparameters.}
As intuition suggests, our detection results provide
strong evidence that context and multi-scale representations improve small
object detection.
\vspace{-1em}
\end{abstract}

%%%%%%%%%%%%%%%%%%%%%%%%%%%%%%%%%%%%%%%%%%%%%%
% Move as needed.
\newcommand{\ERCOCOFig}[1]{\includegraphics[height=0.225\textwidth]{figures/2015test-dev/segclass_rnn_layer2_x3rcnn_conv345normconcat_thresh0_nodrop_coco_ft/#1.png}}

\begin{figure}[t]
  \centering
  \includegraphics[width=0.99\linewidth]{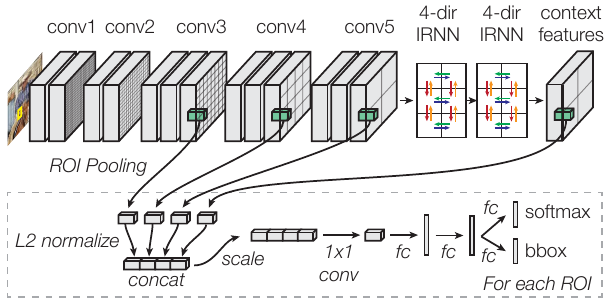}
  \caption{%
    \textbf{Inside-Outside Net (ION).}
    In a single pass, we extract VGG16~\cite{simonyan2015verydeep} features and
    evaluate 2000 proposed regions of interest (ROI).
    For each proposal, we extract a fixed-size
    descriptor from several layers using ROI pooling~\cite{fast-rcnn}.  Each
    descriptor is L2-normalized, concatenated, scaled, and
    dimension-reduced (1x1 convolution) to produce a fixed-length feature
    descriptor for each proposal of size 512x7x7.  Two fully-connected (fc)
    layers process each descriptor and produce two outputs: a one-of-$K$ class
    prediction (``softmax''), and an adjustment to the bounding box (``bbox'').
  }
  \label{fig:arch-roi}

  \vspace{1em}
  \begin{tabular}{@{}c@{}c@{}}
    \ERCOCOFig{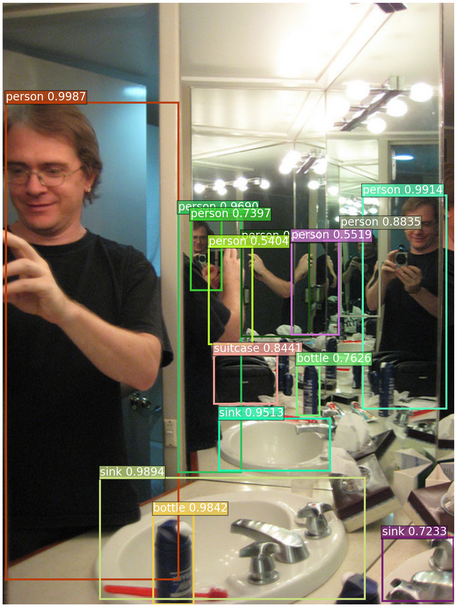}
    \ERCOCOFig{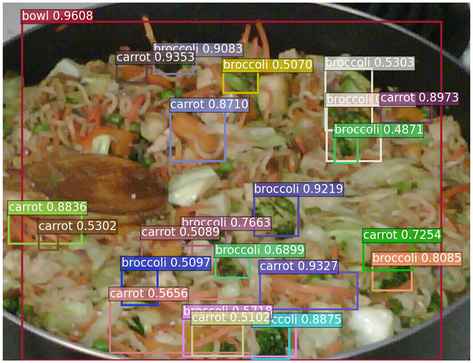} &
    %\ERCOCOFig{32895}
    %\ERCOCOFig{196693}
  \end{tabular}
  %\vspace{-1pt}
  \caption{%
    \textbf{Challenging detections} on COCO 2015 test-dev using our model
    trained on COCO ``2014 train.''
    %\sean{Do we want something like this here?}
  }
  \label{fig:dets}
  \vspace{-6pt}
\end{figure}

%\begin{figure}[t]
%  \begin{center}
%   \includegraphics[width=0.99\linewidth]{figures/arch-roi-small.pdf}
%  \end{center}
%  \vspace{-12pt}
%  \caption{\textbf{ALTERNATE VERSION OF FIGURE~\ref{fig:arch-roi-small}}}
%  \label{fig:arch-roi-small-alt}
%\end{figure}

%%%%%%%%%%%%%%%%%%%%%%%%%%%%%%%%%%%%%%%%%%%%%%

\section{Introduction}

%% START: Larry's text
Reliably detecting an object requires a variety of information, including the
object's fine-grained details and the context surrounding it.
Current state-of-the-art detection approaches~\cite{MR-CNN,ren2015faster} only
use information near an object's region of interest (ROI).  This places
constraints on the type and accuracy of objects that may be detected.

We explore expanding the approach of~\cite{fast-rcnn} to include two additional
sources of information. The first uses a multi-scale
representation~\cite{koenderink1987jets} that captures fine-grained details by
pooling from multiple lower-level convolutional layers in a ConvNet
\cite{simonyan2015verydeep}. These
skip-layers~\cite{sermanetCVPR13,parsenet,fcn,hariharan2015hypercolumns} span
multiple spatial resolutions and levels of feature abstraction. The information
gained is especially important for small objects, which require the higher
spatial resolution provided by lower-level layers.

Our second addition is the use of contextual information. It is well known in
the study of human and computer vision that context plays an important role in
visual recognition~\cite{torralba2003context,deviExploringTinyImages,Hoiem2009}.
To gather contextual information we explore the use of spatial Recurrent Neural Networks (RNNs).
These RNNs pass spatially varying contextual information both horizontally
and vertically across an image. The use of at least two RNN layers ensures
information may be propagated across the entire image. We compare our approach to other common methods for adding contextual information, including global average pooling and additional convolutional layers.
Global average pooling provides information about the entire image, similar to
the features used for scene or image classification
\cite{GIST,krizhevsky2012imagenet}.

Following previous approaches~\cite{girshick2014rcnn}, we use object proposal
detectors \cite{Hosang15proposals,UijlingsIJCV2013,edgeboxes} to identify ROIs in
an image. Each ROI is then classified as containing one or none of the objects of
interest. Using dynamic pooling~\cite{sppnet} we can efficiently evaluate
thousands of different candidate ROIs with a single forwards pass of the
network. For each candidate ROI, the multi-scale and context information is
concatenated into a single layer and fed through several fully connected layers
for classification.

We demonstrate that both sources of additional information, context and
multi-scale, are complementary in nature. This matches our intuition that
context features look broadly across the image, while multi-scale features
capture more fine-grained details. We show large improvements on the PASCAL
VOC~\cite{pascal} and Microsoft COCO~\cite{coco} object detection datasets and
provide a thorough evaluation of the gains across different object types.  We
find that the they are most significant for object types that have been
historically difficult. For example, we show improved accuracy for potted plants
%(\kb{We don't beat MR-CNN on bottle. Sean I think mentioned that plant might be better to mention.}),
which are often small and amongst clutter. In general, we find that our
approach is more adept at detecting small objects than previous
state-of-the-art methods.  For heavily occluded objects like chairs, gains are
found when using contextual information.

While the technical methods employed (spatial
RNNs~\cite{nd-rnn,sceneLSTM,renet}, skip-layer
connections~\cite{sermanetCVPR13,parsenet,fcn,hariharan2015hypercolumns}) have
precedents in the literature, we demonstrate that their well-executed
combination has an unexpectedly positive impact on the detector's accuracy.  As
always, the devil is in the details~\cite{Chatfield14} and thus our paper aims
to provide a thorough exploration of design choices and their
outcomes.

\vspace{-3pt}
\paragraph{Contributions.}
We make the following contributions:
\begin{packed_enum}
\item We introduce the ION architecture that leverages context and multi-scale
  skip pooling for object detection.
%Our architecture uses a 2x stacked 4-directional
%IRNN for context, and multi-layer ROI pooling with feature amplitude
%normalization.

\item We achieve state-of-the-art results on PASCAL VOC 2007, with a mAP of
  \BestVOCSevenTest{}\%, VOC 2012, with a mAP of \BestVOCTwelveTest{}\%,
  and on COCO, with a mAP of \BestCOCOTest{}\%.

\item We conduct extensive experiments evaluating choices like the number of
layers combined, using a segmentation loss, normalizing feature amplitudes,
different IRNN architectures, and other variations.

\item We analyze the detector's performance and find improved accuracy across
the board, but, in particular, for small objects.
%This finding meshes well with
%the intuition behind pursuing context and multi-scale representations.
\end{packed_enum}

%\kb{Discussion: Say something about the per category improvements, small
%objects, other insights. }

\section{Prior work}

\paragraph{ConvNet object detectors.}
ConvNets with a small number of hidden layers have been used for object
detection for the last two decades (\eg, from~\cite{lecun94} to~\cite{sermanetCVPR13}).  Until
recently, they were successful in restricted domains such as face detection.
Recently, deeper ConvNets have led to radical improvements in the
detection of more general object categories.  This shift came about when the
successful application of deep ConvNets to image
classification~\cite{krizhevsky2012imagenet} was transferred to object detection
in the R-CNN system of Girshick \etal~\cite{girshick2014rcnn} and the OverFeat
system of Sermanet \etal~\cite{overfeat}.  Our work builds on the rapidly
evolving R-CNN (``region-based convolutional neural network'') line of work.
Our experiments are conducted with Fast R-CNN~\cite{fast-rcnn},
which is an end-to-end trainable refinement of He \etal's SPPnet~\cite{sppnet}.
We discuss the relationship of our approach to other methods later in the paper
in the context of our model description and experimental results.

\vspace{-3pt}
\paragraph{Spatial RNNs.}
Recurrent Neural Networks (RNNs) exist in various extended forms, including
bidirectional RNNs~\cite{bi-rnn} that process sequences left-to-right and
right-to-left in parallel.  Beyond simple sequences, RNNs exist in full
multi-dimensional variants, such as those introduced by Graves and
Schmidhuber~\cite{nd-rnn} for handwriting recognition.  As a lower-complexity
alternative, \cite{sceneLSTM,renet} explore running an RNN spatially (or
laterally) over a feature map in place of convolutions. These papers examine
spatial RNNs for the tasks of semantic segmentation and image classification,
respectively.  We employ spatial RNNs as a mechanism for computing contextual
features for use in object detection.

\vspace{-3pt}
\paragraph{Skip-layer connections.}
Skip-layer connections are a classic neural network idea wherein activations
from a lower layer are routed directly to a higher layer while bypassing intermediate layers.  The specifics of the wiring and combination method
differ between models and applications.  Our usage of skip connections is most
closely related to those used by Sermanet \etal~\cite{sermanetCVPR13} (termed
``multi-stage features'') for pedestrian detection.  Different
from~\cite{sermanetCVPR13}, we find it essential to L2 normalize activations
from different layers prior to combining them.
%This normalization is necessary
%because our base model, a deep ConvNet that was pre-trained on ImageNet,
%exhibits activations of very different amplitudes across different layers.

The need for activation normalization when combining features across layers was
recently noted by Liu \etal(ParseNet~\cite{parsenet}) in a model for semantic segmentation
that makes use of global image context features.  Skip connections have also
been popular in recent models for semantic segmentation, such as the ``fully
convolutional networks'' in~\cite{fcn}, and for object instance segmentation,
such as the ``hypercolumn features'' in~\cite{hariharan2015hypercolumns}.

%\todo{Point out how past ``Devil in the details'' papers were very useful.}

%%%%%%%%%%%%%%%%%%%%%%%%%%%%%%%%%%%%%%%%%%%%%%
% Move as needed.
\begin{figure*}[t]
  \begin{center}
   \includegraphics[width=0.95\linewidth]{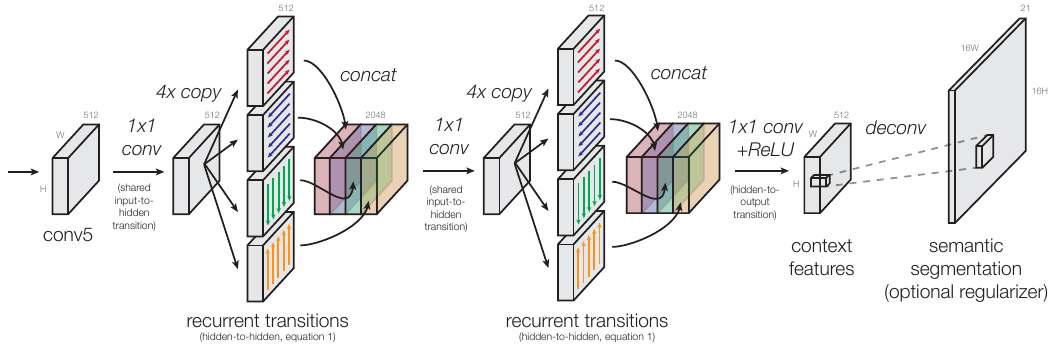}
  \end{center}
  \vspace{-12pt}
  \caption{%
    \textbf{Four-directional IRNN architecture.}
    We use ``IRNN'' units~\cite{irnn} which are RNNs with ReLU recurrent
    transitions, initialized to the identity.  All transitions to/from the
    hidden state are computed with 1x1 convolutions, which allows us to compute
    the recurrence more efficiently (Eq.~\ref{eq:rnn}).  When computing the
    context features, the spatial resolution remains the same throughout (same
    as conv5).  The semantic segmentation regularizer has a 16x higher
    resolution; it is optional and gives a small improvement of around +1 mAP
    point.
  }
  \label{fig:arch-rnn}
\end{figure*}

%%%%%%%%%%%%%%%%%%%%%%%%%%%%%%%%%%%%%%%%%%%%%%

\section{Architecture: Inside-Outside Net (ION)}

In this section we describe ION, a detector with an improved descriptor both
inside and outside the ROI.  An image is processed
by a single deep ConvNet, and the convolutional
feature maps at each stage of the ConvNet are stored in memory. At the top of the network,
a 2x stacked 4-directional IRNN (explained later) computes context features that
describe the image both globally and locally.  The context features have the same dimensions as
``conv5.''  This is done once per image.  In addition, we have thousands of
proposal regions (ROIs) that might contain objects.  For each ROI, we
extract a fixed-length feature descriptor from several layers (``conv3'',
``conv4'', ``conv5'', and ``context features'').  The descriptors are
L2-normalized, concatenated, re-scaled, and dimension-reduced (1x1 convolution)
to produce a fixed-length feature descriptor for each proposal of size 512x7x7.  Two
fully-connected (FC) layers process each descriptor and produce two outputs: a
one-of-$K$ object class prediction (``softmax''), and an adjustment to the proposal
region's bounding box (``bbox'').

The rest of this section explains the details of ION and motivates why we chose
this particular architecture.

%\begin{figure}[t]
%  \begin{center}
%   \includegraphics[width=0.75\linewidth]{figures/naive-roi.pdf}
%  \end{center}
%  \vspace{-12pt}
%  \caption{Naive approach to multi-layer ROI pooling.  This does not work, since
%  the features have very different amplitude.}
%  \label{fig:naive-roi}
%\end{figure}

\subsection{Pooling from multiple layers}
\label{sec:pool-multiple}

% This is redundant given the above:
%As shown in Figure~\ref{fig:arch-roi}, for each proposal region, we pool out of
%several different feature maps and combine all pooled features into a fixed
%output size.  Each of the layers we pool from has a different tradeoff between
%localization, semantic information, and context.  Higher layers have more
%context and semantic information, but carry less information about the precise
%location of objects.

% This is already in prior work:
%Prior methods have used skip connections to improve results in semantic
%segmentation (FCN~\cite{fcn}, ParseNet~\cite{parsenet}), object instance
%segmentation (Hypercolumns~\cite{hariharan2015hypercolumns}), and pedestrian
%detection~\cite{sermanetCVPR13}.  Given their effectiveness, we applied this
%idea of skip connections to object detection via ROI pooling.

Recent successful detectors such as Fast R-CNN, Faster
R-CNN~\cite{ren2015faster}, and SPPnet, all pool from the last convolutional
layer (``conv5\_3'') in VGG16~\cite{simonyan2015verydeep}.  In order to extend
this to multiple layers, we must consider issues of dimensionality and
amplitude.

Since we know that pre-training on ImageNet is important to achieve
state-of-the-art performance~\cite{agrawal14analyzing}, and we would like to use
the previously trained VGG16 network~\cite{simonyan2015verydeep}, it is
important to preserve the existing layer shapes.  Therefore, if we want to pool
out of more layers, the final feature must also be shape 512x7x7 so that it is
the correct shape to feed into the first fully-connected layer (fc6).  In
addition to matching the original shape, we must also match the original
activation amplitudes, so that we can feed our feature into fc6.

To match the required 512x7x7 shape, we concatenate each pooled feature along
the channel axis and reduce the dimension with a 1x1 convolution.  To match the
original amplitudes, we L2 normalize each pooled ROI and re-scale back up by an
empirically determined scale.  Our experiments use a ``scale layer'' with a
learnable per-channel scale initialized to 1000 (measured on the training set).
We later show in Section~\ref{sec:norm} that a fixed scale works just as well.

%To determine what scale to use, we measure the L2 norm of the ROI blobs pooled
%from Fast R-CNN, and then set our scale layer to match it.  We find that the L2
%norm is approximately 1000 on the training set.
%\kb{Given that it does not matter much, is
%it worth discussing it here? Or stick it in supplementary material then?}.

%The concept of combining features from different layers has been widely used in
%prior work including Hypercolumn~\cite{hypercolumn}.  In prior work, different
%layers are typically combined by concatenation, max pooling, summation.  We show
%that for VGG16, this does not work, since the features have a very different
%scale.

As a final note, as more features are concatenated together, we need to
correspondingly decrease the initial weight magnitudes of the 1x1 convolution,
so we use ``Xavier'' initialization~\cite{glorot10}.

\subsection{Context features with IRNNs}

Our architecture for computing context features in ION is shown in more detail in
Figure~\ref{fig:arch-rnn}.  On top of the last convolutional layer (conv5),
we place RNNs that move \emph{laterally} across the image.  Traditionally, an
RNN moves left-to-right along a sequence, consuming an input at every step,
updating its hidden state, and producing an output.  We extend this to two
dimensions by placing an RNN along each row and along each column of the
image.  We have four RNNs in total that move in the cardinal directions: right,
left, down, up.  The RNNs sit on top of conv5 and produce an output with the
same shape as conv5.
%A similar architecture was proposed by ReNet~\cite{ReNet},
%where every layer in the entire network is RNNs moving left/right or up/down.
%This is also similar to the multidimensional\ldots~\cite{grid-rnns}.

There are many possible forms of recurrent neural networks that we could use:
gated recurrent units (GRU)~\cite{gru}, long short-term memory
(LSTM)~\cite{lstm}, and plain tanh recurrent neural networks.  In this paper, we
explore RNNs composed of rectified linear units (ReLU).  Le \etal~\cite{irnn} recently
showed that these networks are easy to train and are good at modeling long-range
dependencies, if the recurrent weight matrix is initialized to the identity
matrix.  This means that at initialization, gradients are propagated backwards
with full strength.
Le \etal~\cite{irnn} call a ReLU RNN initialized this way an ``IRNN,'' and show
that it performs almost as well as an LSTM for a real-world language modeling
task, and better than an LSTM for a toy memory problem.  We adopt this
architecture because it is very simple to implement and parallelize, and is much
faster than LSTMs or GRUs to compute.

For our problem, we have four independent IRNNs that move in four directions.
To implement the IRNNs as efficiently as possible, we split the internal IRNN
computations into separate logical layers.  Viewed this way, we can see that
the input-to-hidden transition is a 1x1 convolution, and that it can be shared
across different directions.
Sharing this transition allows us to remove 6 conv layers in total with a
negligible effect on accuracy ($-0.1$ mAP).
The bias can be shared in the same way, and
merged into the 1x1 conv layer.  The IRNN layer now only needs to apply
the recurrent matrix and apply the nonlinearity at each step.  The output
from the IRNN is computed by concatenating the hidden state from the four
directions at each spatial location.

%With this setup, the IRNN layer can do an in-place accumulation.
This is the update for an IRNN that moves to the right; similar equations exist
for the other directions:
\begin{align}
  h_{i,j}^\text{right} &\gets \max\left(
    \mathbf{W}_{hh}^\text{right} h_{i,j-1}^\text{right} + h_{i,j}^\text{right},
    0 \right).
  \label{eq:rnn}
\end{align}
Notice that the input is not explicitly shown in the equation, and there is no
input-to-hidden transition.  This is because it was computed as part of the 1x1
convolution, and then copied in-place to each hidden layer.
For each direction, we can compute all of the independent rows/columns in
parallel, stepping all IRNNs together with a single matrix multiply.  On a GPU,
this results in large speedups compared to computing each RNN cell one at a
time.
%\kb{Can you characterize how much? Ballpark?}
% ^ Sean: No, I never compared directly, but I know it would be a large
% difference.

We also explore using semantic segmentation labels to regularize the IRNN
output.  When using these labels, we add the deconvolution and crop layer as
implemented by Long \etal~\cite{fcn}.  The deconvolution upsamples by 16x with a
32x32 kernel, and we add an extra softmax loss layer with a weight of 1.  This
is evaluated in Section~\ref{sec:seg}.

\paragraph{Variants and simplifications.} We explore several
further simplifications.

\begin{packed_enum}
\item We fixed the hidden transition matrix to the identity $\mathbf{W}_{hh}^\text{right}
= I$, which allows us to entirely remove it:
\begin{align}
  h_{i,j}^\text{right} &\gets \max\left( h_{i,j-1}^\text{right} + h_{i,j}^\text{right}, 0 \right).
  \label{eq:no-whh}
\end{align}
This is like an accumulator, but with ReLU after each step.  In
Section~\ref{sec:irnn-arch} we show that removing the recurrent matrix has a
surprisingly small impact.
\item To prevent overfitting, we include dropout layers ($p = 0.25$)
after each concat layer in all experiments.  We later found that in fact the model is
underfitting and there is no need for dropout anywhere in the network.
\item Finally, we trained a separate bias $b_0$ for the first step in the RNN
in each direction. However, since it tends to remain near zero after training,
this component is not really necessary.
\end{packed_enum}

\begin{figure}[t]
  \begin{center}
   \includegraphics[width=0.6\linewidth]{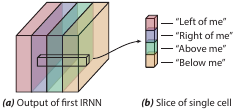}
  \end{center}
  \vspace{-6pt}
  \caption{Interpretation of the first IRNN output.  Each cell in the output
  summarizes the features to the left/right/top/bottom.}
  \label{fig:rnn-slice}
  \vspace{-6pt}
\end{figure}

\paragraph{Interpretation.}
After the first 4-directional IRNN (out of the two IRNNs), we obtain a feature map that summarizes
nearby objects at every position in the image.  As illustrated in
Figure~\ref{fig:rnn-slice}, we can see that the first IRNN creates a summary of
the features to the left/right/top/bottom of every cell.  The subsequent 1x1
convolution then mixes this information together as a dimension reduction.

After the second 4-directional IRNN, every cell on the output depends on every
cell of the input.  In this way, our context features are both global and
local.  The features vary by spatial position, and each cell is a global
summary of the image with respect to that specific spatial location.

% merge all three tabulars into one table environment, with separate captions,
% so that we can force them to be together and control spacing exactly
\begin{table*}[t]
  \centering
    \footnotesize{%
    \begin{tabular}{%
        @{\hskip 0.2em}p{2.2cm}
        @{\hskip 0.2em}c
        @{\hskip 0.2em}c
        @{\hskip 0.2em}c
        @{\hskip 0.2em}c@{\hskip 0.5em}|
        @{\hskip 0.5em}l@{\hskip 0.5em}|
        @{\hskip 0.5em}c@{\hskip 0.5em}|
        @{\hskip 0.5em}c @{\hskip 0.5em}c @{\hskip 0.5em}c @{\hskip 0.5em}c
        @{\hskip 0.5em}c @{\hskip 0.5em}c @{\hskip 0.5em}c @{\hskip 0.5em}c
        @{\hskip 0.5em}c @{\hskip 0.5em}c @{\hskip 0.5em}c @{\hskip 0.5em}c
        @{\hskip 0.5em}c @{\hskip 0.5em}c @{\hskip 0.5em}c @{\hskip 0.5em}c
        @{\hskip 0.5em}c @{\hskip 0.5em}c @{\hskip 0.5em}c @{\hskip 0.5em}c
        @{\hskip 0.2em}
      }
      \toprule
      \textbf{Method} &
      \textbf{R} &
      \textbf{S} &
      \textbf{W} &
      \textbf{D} &
      \textbf{Train} &
      \textbf{mAP} &
      \tiny{\textbf{aero}} &
      \tiny{\textbf{bike}} &
      \tiny{\textbf{bird}} &
      \tiny{\textbf{boat}} &
      \tiny{\textbf{bottle}} &
      \tiny{\textbf{bus}} &
      \tiny{\textbf{car}} &
      \tiny{\textbf{cat}} &
      \tiny{\textbf{chair}} &
      \tiny{\textbf{cow}} &
      \tiny{\textbf{table}} &
      \tiny{\textbf{dog}} &
      \tiny{\textbf{horse}} &
      \tiny{\textbf{mbike}} &
      \tiny{\textbf{person}} &
      \tiny{\textbf{plant}} &
      \tiny{\textbf{sheep}} &
      \tiny{\textbf{sofa}} &
      \tiny{\textbf{train}} &
      \tiny{\textbf{tv}} \\
      \midrule
      FRCN~\cite{fast-rcnn} & & & & & 07+12 &
      70.0 &
      \scriptsize{77.0} &
      \scriptsize{78.1} &
      \scriptsize{69.3} &
      \scriptsize{59.4} &
      \scriptsize{38.3} &
      \scriptsize{81.6} &
      \scriptsize{78.6} &
      \scriptsize{86.7} &
      \scriptsize{42.8} &
      \scriptsize{78.8} &
      \scriptsize{68.9} &
      \scriptsize{84.7} &
      \scriptsize{82.0} &
      \scriptsize{76.6} &
      \scriptsize{69.9} &
      \scriptsize{31.8} &
      \scriptsize{70.1} &
      \scriptsize{74.8} &
      \scriptsize{80.4} &
      \scriptsize{70.4}
      \\
      %FRCN*~\cite{fast-rcnn} & & & & & 07+12 &
      %70.8 &
      %\scriptsize{75.9} &
      %\scriptsize{78.3} &
      %\scriptsize{70.8} &
      %\scriptsize{58.8} &
      %\scriptsize{46.5} &
      %\scriptsize{82.2} &
      %\scriptsize{77.9} &
      %\scriptsize{85.0} &
      %\scriptsize{49.0} &
      %\scriptsize{75.2} &
      %\scriptsize{66.3} &
      %\scriptsize{83.4} &
      %\scriptsize{81.9} &
      %\scriptsize{76.3} &
      %\scriptsize{69.3} &
      %\scriptsize{39.4} &
      %\scriptsize{70.6} &
      %\scriptsize{72.6} &
      %\scriptsize{79.4} &
      %\scriptsize{76.7}
      %\\
      %FRCN*~\cite{fast-rcnn} & & & \checkmark & & 07+12 &
      %74.6 &
      %\scriptsize{76.9} &
      %\scriptsize{81.4} &
      %\scriptsize{73.2} &
      %\scriptsize{66.4} &
      %\scriptsize{52.8} &
      %\scriptsize{83.4} &
      %\scriptsize{82.1} &
      %\scriptsize{86.7} &
      %\scriptsize{53.6} &
      %\scriptsize{82.1} &
      %\scriptsize{71.0} &
      %\scriptsize{86.7} &
      %\scriptsize{86.7} &
      %\scriptsize{78.0} &
      %\scriptsize{75.0} &
      %\scriptsize{44.2} &
      %\scriptsize{75.6} &
      %\scriptsize{73.1} &
      %\scriptsize{83.4} &
      %\scriptsize{79.6}
      %\\
      %\midrule
      RPN~\cite{ren2015faster} & & & & & 07+12 &
      {73.2} &
      \scriptsize{76.5} &
      \scriptsize{79.0} &
      \scriptsize{70.9} &
      \scriptsize{65.5} &
      \scriptsize{52.1} &
      \scriptsize{83.1} &
      \scriptsize{84.7} &
      \scriptsize{86.4} &
      \scriptsize{52.0} &
      \scriptsize{81.9} &
      \scriptsize{65.7} &
      \scriptsize{84.8} &
      \scriptsize{84.6} &
      \scriptsize{77.5} &
      \scriptsize{76.7} &
      \scriptsize{38.8} &
      \scriptsize{73.6} &
      \scriptsize{73.9} &
      \scriptsize{83.0} &
      \scriptsize{72.6}
      \\
      MR-CNN~\cite{MR-CNN} & & & \checkmark & & 07+12 &
      {78.2} &
      \scriptsize{\textbf{80.3}} &
      \scriptsize{84.1} &
      \scriptsize{78.5} &
      \scriptsize{70.8} &
      \scriptsize{\textbf{68.5}} &
      \scriptsize{\textbf{88.0}} &
      \scriptsize{85.9} &
      \scriptsize{87.8} &
      \scriptsize{60.3} &
      \scriptsize{85.2} &
      \scriptsize{73.7} &
      \scriptsize{87.2} &
      \scriptsize{86.5} &
      \scriptsize{\textbf{85.0}} &
      \scriptsize{76.4} &
      \scriptsize{48.5} &
      \scriptsize{76.3} &
      \scriptsize{75.5} &
      \scriptsize{85.0} &
      \scriptsize{81.0}
      \\
      \midrule
      ION [ours] & & & & & 07+12 &
      {74.6} &
      \scriptsize{78.2} &
      \scriptsize{79.1} &
      \scriptsize{76.8} &
      \scriptsize{61.5} &
      \scriptsize{54.7} &
      \scriptsize{81.9} &
      \scriptsize{84.3} &
      \scriptsize{88.3} &
      \scriptsize{53.1} &
      \scriptsize{78.3} &
      \scriptsize{71.6} &
      \scriptsize{85.9} &
      \scriptsize{84.8} &
      \scriptsize{81.6} &
      \scriptsize{74.3} &
      \scriptsize{45.6} &
      \scriptsize{75.3} &
      \scriptsize{72.1} &
      \scriptsize{82.6} &
      \scriptsize{81.4}
      \\
      ION [ours] & \checkmark & & & & 07+12 &
      {75.6} &
      \scriptsize{79.2} &
      \scriptsize{83.1} &
      \scriptsize{77.6} &
      \scriptsize{65.6} &
      \scriptsize{54.9} &
      \scriptsize{85.4} &
      \scriptsize{85.1} &
      \scriptsize{87.0} &
      \scriptsize{54.4} &
      \scriptsize{80.6} &
      \scriptsize{73.8} &
      \scriptsize{85.3} &
      \scriptsize{82.2} &
      \scriptsize{82.2} &
      \scriptsize{74.4} &
      \scriptsize{47.1} &
      \scriptsize{75.8} &
      \scriptsize{72.7} &
      \scriptsize{84.2} &
      \scriptsize{80.4}
      \\
      ION [ours] & \checkmark & \checkmark & & & 07+12+S &
      {76.5} &
      \scriptsize{79.2} &
      \scriptsize{79.2} &
      \scriptsize{77.4} &
      \scriptsize{69.8} &
      \scriptsize{55.7} &
      \scriptsize{85.2} &
      \scriptsize{84.2} &
      \scriptsize{89.8} &
      \scriptsize{57.5} &
      \scriptsize{78.5} &
      \scriptsize{73.8} &
      \scriptsize{87.8} &
      \scriptsize{85.9} &
      \scriptsize{81.3} &
      \scriptsize{75.3} &
      \scriptsize{49.7} &
      \scriptsize{76.9} &
      \scriptsize{74.6} &
      \scriptsize{85.2} &
      \scriptsize{82.1}
      \\
      ION [ours] & \checkmark & \checkmark & \checkmark & & 07+12+S &
      {78.5} &
      \scriptsize{{80.2}} &
      \scriptsize{84.7} &
      \scriptsize{\textbf{78.8}} &
      \scriptsize{\textbf{72.4}} &
      \scriptsize{61.9} &
      \scriptsize{86.2} &
      \scriptsize{86.7} &
      \scriptsize{89.5} &
      \scriptsize{59.1} &
      \scriptsize{84.1} &
      \scriptsize{74.7} &
      \scriptsize{\textbf{88.9}} &
      \scriptsize{86.9} &
      \scriptsize{81.3} &
      \scriptsize{80.0} &
      \scriptsize{50.9} &
      \scriptsize{\textbf{80.4}} &
      \scriptsize{74.1} &
      \scriptsize{86.6} &
      \scriptsize{83.3}
      \\
      ION [ours] & \checkmark & \checkmark & \checkmark & \checkmark & 07+12+S &
      \textbf{79.2} &
      \scriptsize{{80.2}} &
      \scriptsize{\textbf{85.2}} &
      \scriptsize{\textbf{78.8}} &
      \scriptsize{70.9} &
      \scriptsize{{62.6}} &
      \scriptsize{{86.6}} &
      \scriptsize{\textbf{86.9}} &
      \scriptsize{\textbf{89.8}} &
      \scriptsize{\textbf{61.7}} &
      \scriptsize{\textbf{86.9}} &
      \scriptsize{\textbf{76.5}} &
      \scriptsize{88.4} &
      \scriptsize{\textbf{87.5}} &
      \scriptsize{{83.4}} &
      \scriptsize{\textbf{80.5}} &
      \scriptsize{\textbf{52.4}} &
      \scriptsize{78.1} &
      \scriptsize{\textbf{77.2}} &
      \scriptsize{\textbf{86.9}} &
      \scriptsize{\textbf{83.5}}
      \\
      \bottomrule
    \end{tabular}
  }
  \caption{%
    \textbf{Detection results on VOC 2007 test.}
    Legend:
    \textbf{07+12:} 07 trainval + 12 trainval,
    \textbf{07+12+S:} 07+12 plus SBD segmentation labels~\cite{BharathICCV2011},
    \textbf{R:} include 2x stacked 4-dir IRNN (context features),
    \textbf{S:} regularize with segmentation labels,
    \textbf{W:} two rounds of bounding box regression and weighted voting~\cite{MR-CNN},
    \textbf{D:} remove all dropout layers.
  }
  \label{tab:voc-2007-test}

  \vspace{1.25em}
    \footnotesize{%
    \begin{tabular}{%
        @{\hskip 0.2em}p{2.2cm}
        @{\hskip 0.2em}c
        @{\hskip 0.2em}c
        @{\hskip 0.2em}c
        @{\hskip 0.2em}c@{\hskip 0.5em}|
        @{\hskip 0.5em}l@{\hskip 0.5em}|
        @{\hskip 0.5em}c@{\hskip 0.5em}|
        @{\hskip 0.5em}c @{\hskip 0.5em}c @{\hskip 0.5em}c @{\hskip 0.5em}c
        @{\hskip 0.5em}c @{\hskip 0.5em}c @{\hskip 0.5em}c @{\hskip 0.5em}c
        @{\hskip 0.5em}c @{\hskip 0.5em}c @{\hskip 0.5em}c @{\hskip 0.5em}c
        @{\hskip 0.5em}c @{\hskip 0.5em}c @{\hskip 0.5em}c @{\hskip 0.5em}c
        @{\hskip 0.5em}c @{\hskip 0.5em}c @{\hskip 0.5em}c @{\hskip 0.5em}c
        @{\hskip 0.2em}
      }
      \toprule
      \textbf{Method} &
      \textbf{R} &
      \textbf{S} &
      \textbf{W} &
      \textbf{D} &
      \textbf{Train} &
      \textbf{mAP} &
      \tiny{\textbf{aero}} &
      \tiny{\textbf{bike}} &
      \tiny{\textbf{bird}} &
      \tiny{\textbf{boat}} &
      \tiny{\textbf{bottle}} &
      \tiny{\textbf{bus}} &
      \tiny{\textbf{car}} &
      \tiny{\textbf{cat}} &
      \tiny{\textbf{chair}} &
      \tiny{\textbf{cow}} &
      \tiny{\textbf{table}} &
      \tiny{\textbf{dog}} &
      \tiny{\textbf{horse}} &
      \tiny{\textbf{mbike}} &
      \tiny{\textbf{person}} &
      \tiny{\textbf{plant}} &
      \tiny{\textbf{sheep}} &
      \tiny{\textbf{sofa}} &
      \tiny{\textbf{train}} &
      \tiny{\textbf{tv}} \\
      \midrule
      FRCN~\cite{fast-rcnn} & & & & & 07++12 &
      68.4 &
      \scriptsize{82.3} &
      \scriptsize{78.4} &
      \scriptsize{70.8} &
      \scriptsize{52.3} &
      \scriptsize{38.7} &
      \scriptsize{77.8} &
      \scriptsize{71.6} &
      \scriptsize{89.3} &
      \scriptsize{44.2} &
      \scriptsize{73.0} &
      \scriptsize{55.0} &
      \scriptsize{87.5} &
      \scriptsize{80.5} &
      \scriptsize{80.8} &
      \scriptsize{72.0} &
      \scriptsize{35.1} &
      \scriptsize{68.3} &
      \scriptsize{65.7} &
      \scriptsize{80.4} &
      \scriptsize{64.2}
      \\
      RPN~\cite{ren2015faster} & & & & & 07++12 &
      70.4 &
      \scriptsize{84.9} &
      \scriptsize{79.8} &
      \scriptsize{74.3} &
      \scriptsize{53.9} &
      \scriptsize{49.8} &
      \scriptsize{77.5} &
      \scriptsize{75.9} &
      \scriptsize{88.5} &
      \scriptsize{45.6} &
      \scriptsize{77.1} &
      \scriptsize{55.3} &
      \scriptsize{86.9} &
      \scriptsize{81.7} &
      \scriptsize{80.9} &
      \scriptsize{79.6} &
      \scriptsize{40.1} &
      \scriptsize{72.6} &
      \scriptsize{60.9} &
      \scriptsize{81.2} &
      \scriptsize{61.5}
      \\
      FRCN+YOLO~\cite{yolo} & & & & & 07++12 &
      70.4 &
      \scriptsize{83.0} &
      \scriptsize{78.5} &
      \scriptsize{73.7} &
      \scriptsize{55.8} &
      \scriptsize{43.1} &
      \scriptsize{78.3} &
      \scriptsize{73.0} &
      \scriptsize{89.2} &
      \scriptsize{49.1} &
      \scriptsize{74.3} &
      \scriptsize{56.6} &
      \scriptsize{87.2} &
      \scriptsize{80.5} &
      \scriptsize{80.5} &
      \scriptsize{74.7} &
      \scriptsize{42.1} &
      \scriptsize{70.8} &
      \scriptsize{68.3} &
      \scriptsize{81.5} &
      \scriptsize{67.0}
      \\
      HyperNet & & & & & 07++12 &
      71.4 &
      \scriptsize{84.2} &
      \scriptsize{78.5} &
      \scriptsize{73.6} &
      \scriptsize{55.6} &
      \scriptsize{53.7} &
      \scriptsize{78.7} &
      \scriptsize{79.8} &
      \scriptsize{87.7} &
      \scriptsize{49.6} &
      \scriptsize{74.9} &
      \scriptsize{52.1} &
      \scriptsize{86.0} &
      \scriptsize{81.7} &
      \scriptsize{83.3} &
      \scriptsize{81.8} &
      \scriptsize{48.6} &
      \scriptsize{73.5} &
      \scriptsize{59.4} &
      \scriptsize{79.9} &
      \scriptsize{65.7}
      \\
      MR-CNN~\cite{MR-CNN} & & & \checkmark & & 07+12 &
      73.9 &
      \scriptsize{85.5} &
      \scriptsize{82.9} &
      \scriptsize{76.6} &
      \scriptsize{57.8} &
      \scriptsize{\textbf{62.7}} &
      \scriptsize{79.4} &
      \scriptsize{77.2} &
      \scriptsize{86.6} &
      \scriptsize{55.0} &
      \scriptsize{79.1} &
      \scriptsize{62.2} &
      \scriptsize{87.0} &
      \scriptsize{83.4} &
      \scriptsize{\textbf{84.7}} &
      \scriptsize{78.9} &
      \scriptsize{45.3} &
      \scriptsize{73.4} &
      \scriptsize{65.8} &
      \scriptsize{80.3} &
      \scriptsize{\textbf{74.0}}
      \\
      \midrule
      %ION [ours] & & & & & 2k & &
      %\\
      %ION [ours] & \checkmark & & & & 2k &
      %&
      %\\
      %ION [ours] & \checkmark & \checkmark & & & 2k &
      %&
      %\\
      %ION [ours] & \checkmark & \checkmark & \checkmark & & 2k &
      %&
      %\\
      ION [ours] & \checkmark & \checkmark & \checkmark & \checkmark & 07+12+S &
      \textbf{76.4} &
      \scriptsize{\textbf{87.5}} &
      \scriptsize{\textbf{84.7}} &
      \scriptsize{\textbf{76.8}} &
      \scriptsize{\textbf{63.8}} &
      \scriptsize{58.3} &
      \scriptsize{\textbf{82.6}} &
      \scriptsize{\textbf{79.0}} &
      \scriptsize{\textbf{90.9}} &
      \scriptsize{\textbf{57.8}} &
      \scriptsize{\textbf{82.0}} &
      \scriptsize{\textbf{64.7}} &
      \scriptsize{\textbf{88.9}} &
      \scriptsize{\textbf{86.5}} &
      \scriptsize{\textbf{84.7}} &
      \scriptsize{\textbf{82.3}} &
      \scriptsize{\textbf{51.4}} &
      \scriptsize{\textbf{78.2}} &
      \scriptsize{\textbf{69.2}} &
      \scriptsize{\textbf{85.2}} &
      \scriptsize{73.5}
      \\
      \bottomrule
    \end{tabular}
  }
  \vspace{0pt}
  \caption{%
    \textbf{Detection results on VOC 2012 test (comp4).}
    Legend:
    \textbf{07+12:} 07 trainval + 12 trainval,
    \textbf{07++12:} 07 trainvaltest + 12 trainval,
    \textbf{07+12+S:} 07+12 plus SBD segmentation labels~\cite{BharathICCV2011},
    \textbf{R:} include 2x stacked 4-dir IRNN (context features),
    \textbf{S:} regularize with segmentation labels,
    \textbf{W:} two rounds of bounding box regression and weighted voting~\cite{MR-CNN},
    \textbf{D:} remove all dropout layers.
  }
  \label{tab:voc-2012-test}

  \vspace{1.25em}
    \footnotesize{%
    \begin{tabular}{%
        p{1.6cm}c@{\hskip 0em}c@{\hskip 0.2em}c@{\hskip 0.2em}c
        | l |
        rrr |
        rrr |
        rrr |
        rrr
      }
      \toprule
      \multirow{2}{*}{\textbf{Method}} &
      \multirow{2}{*}{\textbf{R}} &
      \multirow{2}{*}{\textbf{S}} &
      \multirow{2}{*}{\textbf{W}} &
      \multirow{2}{*}{\textbf{D}} &
      \multirow{2}{*}{\textbf{Train}} &
      \multicolumn{3}{|c}{\textbf{Avg. Precision, IoU:}} &
      \multicolumn{3}{|c}{\textbf{Avg. Precision, Area:}} &
      \multicolumn{3}{|c}{\textbf{Avg. Recall, \# Dets:}} &
      \multicolumn{3}{|c}{\textbf{Avg. Recall, Area:}}
      \\
       & & & & & &
       0.5:0.95 & 0.50 & 0.75 &
       Small & Med. & Large &
       1 & 10 & 100 &
       Small & Med. & Large
      \\
      \midrule
      % test-dev for noseg_conv5_thresh0_coco.json.zip
      FRCN~\cite{fast-rcnn}* & & & & & train &
      20.5  &
      39.9  &
      19.4  &
       4.1  &
      20.0  &
      35.8  &
      21.3  &
      29.5  &
      30.1  &
       7.3  &
      32.1  &
      52.0
      \\
      % test-dev for noseg_conv5_thresh0_bbreg2_bbthresh01abs_coco.json.zip
      FRCN~\cite{fast-rcnn}* & & & \checkmark & & train &
      20.0  &
      40.3  &
      18.1  &
       4.1  &
      19.6  &
      34.5  &
      20.8  &
      29.1  &
      29.8  &
       7.4  &
      31.9  &
      50.9
      \\
      \midrule
      % test-dev for detections_test-dev2015_noseg_rnn_layer2_x3rcnn_conv345normconcat_thresh0_coco_ft_iter320000_results.json.zip
      ION [ours] & \checkmark & & & & train &
      23.0 &
      42.0 &
      23.0 &
       6.0 &
      23.8 &
      37.3 &
      23.0 &
      32.4 &
      33.0 &
       9.7 &
      37.0 &
      53.5
      \\
      % test-dev for detections_test-dev2015_noseg_rnn_layer2_x3rcnn_conv345normconcat_thresh0_nodrop_coco_ft_iter320000_results.json.zip
      ION [ours] & \checkmark & & & \checkmark & train &
      23.6 &
      43.2 &
      23.6 &
       6.4 &
      24.1 &
      38.3 &
      23.2 &
      32.7 &
      33.5 &
      10.1 &
      37.7 &
      53.6
      %\\
      %% test-dev for detections_test-dev2015_rnn_layer2_x3rcnn_conv345normconcat_thresh0_coco_ft_iter220000_results.json.zip
      %ION [ours] & \checkmark & \checkmark & & & train &
      %24.0 &
      %43.8 &
      %24.0 &
      % 6.6 &
      %24.6 &
      %39.1 &
      %23.4 &
      %32.9 &
      %33.6 &
      %10.2 &
      %37.8 &
      %53.8
      \\
        % test-dev for detections_test-dev2015_rnn_layer2_x3rcnn_conv345normconcat_thresh0_nodrop_coco_ft_iter320000_results.json.
      ION [ours] & \checkmark & \checkmark & & \checkmark & train &
      \newtext{24.9} &
      \newtext{44.7} &
      \newtext{25.3} &
      \newtext{7.0} &
      \newtext{26.1} &
      \newtext{40.1} &
      \newtext{23.9} &
      \newtext{33.5} &
      \newtext{34.1} &
      \newtext{10.7} &
      \newtext{38.8} &
      \newtext{54.1}
      \\
      \midrule
      % test-dev for
      % detections_test-dev2015_conv3457norm_irnn2_seg_sched6_coco100kother_mix1_nrep_seed10_coco_ft-test_mix3_nms443_2xbbregwvall854_dedup8_coco_iter80000_results.json.zip
      \newtext{ION comp.$^\dagger$} & \checkmark & \checkmark &
      \checkmark & \checkmark & trainval35k &
      \newtext{31.2} &
      \newtext{53.4} &
      \newtext{32.3} &
      \newtext{12.8} &
      \newtext{32.9} &
      \newtext{45.2} &
      \newtext{27.8} &
      \newtext{43.1} &
      \newtext{45.6} &
      \newtext{23.6} &
      \newtext{50.0} &
      \newtext{63.2}
      \\
      %% test-dev for
      %% detections_test-dev2015_conv3457norm_irnn2_seg_sched8_coco80kother_mix1_nrep_seed12_coco_ft-test_mix5_nms443_2xbbregwvall854_dedup8_flipens_coco_iter80000_results.json.zip
      %\newtext{ION [ours] post-comp.$^\dagger$} & \checkmark & \checkmark &
      %\checkmark & \checkmark & trainval35k$^\dagger$ &
      %\newtext{\textbf{32.3}} &
      %\newtext{\textbf{54.7}} &
      %\newtext{\textbf{33.7}} &
      %\newtext{\textbf{13.9}} &
      %\newtext{\textbf{34.2}} &
      %\newtext{\textbf{46.4}} &
      %\newtext{\textbf{28.5}} &
      %\newtext{\textbf{44.2}} &
      %\newtext{\textbf{46.8}} &
      %\newtext{\textbf{24.9}} &
      %\newtext{\textbf{51.6}} &
      %\newtext{\textbf{64.0}}
      %\\
      % test-dev for
      % detections_test-dev2015_conv3457norm_irnn2_ft1_seg_sched10_coco80kother_mix1_nrep_seed12_coco_ft-test_mix5_nms443_2xbbregwvall854_dedup8_flipens_coco_iter160000_results.json.zip
      \newtext{ION post.$^\dagger$} & \checkmark & \checkmark &
      \checkmark & \checkmark & trainval35k &
      \newtext{\textbf{33.1}} &
      \newtext{\textbf{55.7}} &
      \newtext{\textbf{34.6}} &
      \newtext{\textbf{14.5}} &
      \newtext{\textbf{35.2}} &
      \newtext{\textbf{47.2}} &
      \newtext{\textbf{28.9}} &
      \newtext{\textbf{44.8}} &
      \newtext{\textbf{47.4}} &
      \newtext{\textbf{25.5}} &
      \newtext{\textbf{52.4}} &
      \newtext{\textbf{64.3}}
      \\
      \bottomrule
    \end{tabular}
  }
  \vspace{0pt}
  \caption{%
    \textbf{Detection results on COCO 2015 test-dev.}
    Legend:
    \textbf{R:} include 2x stacked 4-dir IRNN (context features),
    \textbf{S:} regularize with segmentation labels,
    \textbf{W:} two rounds of bounding box regression and weighted
    voting~\cite{MR-CNN},
    \textbf{D:} remove all dropout layers.
    *We use a longer training schedule, resulting in a higher score than the
    preliminary numbers in~\cite{fast-rcnn}.
    $^\dagger$test-dev scores for our submission to the 2015 MS COCO Detection
    competition, and post-competition improvements, trained on ``trainval35k'', described in the Appendix.
  }
  \label{tab:coco-2015-test}

  \vspace{-3pt}
\end{table*}

\section{Results}

We train and evaluate our dataset on three major datasets: PASCAL VOC 2007, VOC
2012, and on MS COCO.  We demonstrate state-of-the-art results on all three
datasets.

\subsection{Experimental setup}
\label{sec:experimental-setup}

All of our experiments use Fast R-CNN~\cite{fast-rcnn} built on the
Caffe~\cite{jia2014caffe} framework, and the VGG16
architecture~\cite{simonyan2015verydeep}, all of which are available online.
As is common practice, we use the publicly available weights pre-trained on
ILSVRC2012~\cite{ilsvrc2012} downloaded from the Caffe Model
Zoo.\footnote{\scriptsize{\url{https://github.com/BVLC/caffe/wiki/Model-Zoo}}}

We make some changes to Fast R-CNN, which give a small improvement over the
baseline.  We use 4 images per mini-batch, implemented as 4 forward/backward
passes of single image mini-batches, with gradient accumulation.  We sample 128
ROIs per image leading to 512 ROIs per model update.  We measure the norm of the
parameter gradient vector and rescale it if its L2 norm is above 20 (80 when
accumulating over 4 images).

To accelerate training, we use a two-stage schedule.  As noted by
Girshick~\cite{fast-rcnn}, it is not necessary to fine-tune all layers, and
nearly the same performance can be achieved by fine-tuning starting from
conv3\_1.  With this in mind, we first train for {40k} iterations with
conv1\_1 through conv5\_3 frozen, and then another {100k} iterations with
only conv1\_1 through conv2\_2 frozen.  All other layers are fine-tuned.  When training for COCO, we use 80k
and 320k iterations respectively.  We found that shorter training schedules
are not enough to fully converge.

We also use a different learning rate (LR) schedule.  The LR exponentially
decays from $5 \cdot 10^{-3}$ to $10^{-4}$ in the first stage, and from
$10^{-3}$ to $10^{-5}$ in the second stage.  To reduce the effect of random
variation, we fix the random seed so that all variants see the same images in
the same order.  For PASCAL VOC we use the same pre-computed selective search
boxes from Fast R-CNN, and for COCO we use the boxes precomputed by Hosang
\etal~\cite{Hosang15proposals}.  Finally, we modified the test thresholds in
Fast R-CNN so that we keep only boxes with a softmax score above 0.05, and keep
at most 100 boxes per images.

When re-running the baseline Fast R-CNN using the above settings, we see a +0.8
mAP improvement over the original settings on VOC 2007 test.  We compare
against the baseline using our improved settings where possible.

%%%%%%%%%%%%%%%%%%%%%%%%%%%%%%%%%%%%%%%%%%%%%%
% Move as needed.
\begin{figure}[t]
  \begin{center}
   \scriptsize{\textbf{Extra-small Objects}}
   \includegraphics[width=0.95\linewidth]{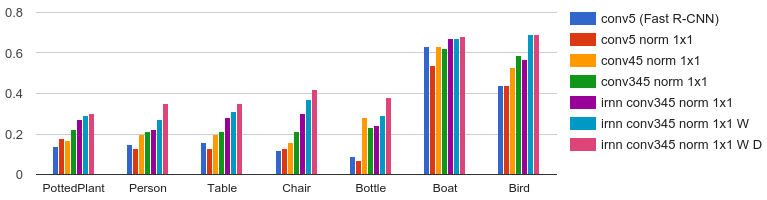} \\
   \scriptsize{\textbf{Medium Objects}}
   \includegraphics[width=0.95\linewidth]{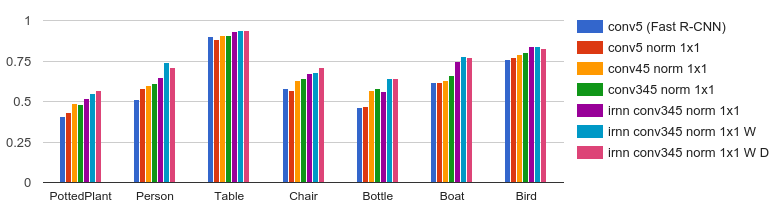} \\
   \scriptsize{\textbf{Extra-large Objects}}
   \includegraphics[width=0.95\linewidth]{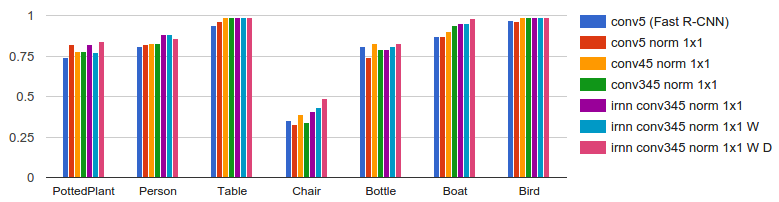}
  \end{center}
  \vspace{-12pt}
  \caption{%
    \textbf{VOC 2007 normalized AP by size.}
    Left to right: increasing complexity.
    Left-most bar in each group: Fast R-CNN; right-most bar: our best model that
    achieves 79.2\% mAP on VOC 2007 test.  Our detector has a particularly
    large improvement for small objects. See Hoiem~\cite{hoiem12error}
for details on these metrics.
    %\textbf{bbreg2wv:} voting scheme from \cite{MR-CNN}.
  }
  \label{fig:xs-objects}
\end{figure}

%%%%%%%%%%%%%%%%%%%%%%%%%%%%%%%%%%%%%%%%%%%%%%

\subsection{PASCAL VOC 2007}

As shown in Table~\ref{tab:voc-2007-test}, we evaluate our detector (ION) on PASCAL
VOC 2007, training on the VOC 2007 trainval dataset merged with the 2012
trainval dataset, a common practice.  Applying our method described above, we obtain a mAP of
76.5\%.  We then make some simple modifications, as described below, to achieve
a higher score of 79.2\%.

MR-CNN~\cite{MR-CNN} introduces a bounding box regression scheme to improve
results on VOC, where bounding boxes are evaluated twice: (1) the initial
proposal boxes are evaluated and regressed to improved locations and then (2)
the improved locations are passed again through the network.  All boxes are
accumulated together, and non-max supression is applied.  Finally, a weighted
vote is computed for each kept box (over all boxes, including those
suppressed), where boxes that overlap a kept box by at least 0.5 IoU contribute
to the average.  For our method, we use the softmax scores as the weights.
When adding this scheme to our method, our mAP rises from 76.5\% to 78.5\%.
Finally, we observed that our models are underfitting and we remove dropout from
all layers to get a further gain up to 79.2\%.

MR-CNN also uses context and achieves 78.2\%.  However, we note that their
method requires that pieces are evaluated individually, and thus has a test
runtime around 30 seconds per image, while our method is significantly faster,
taking 0.8s per image on a Titan X GPU (excluding proposal generation) without
two-stage bounding box regression and 1.15s per image with it.
%\footnote{Our
%implementation is inefficient and can be significantly accelerated.}

\subsection{PASCAL VOC 2012}

We also evaluate on the slightly more challenging VOC 2012 dataset, submitting
to the public evaluation server.\footnote{Anonymous URL:
\tiny{\url{http://host.robots.ox.ac.uk:8080/anonymous/B3VFLE.html}}}
In Table~\ref{tab:voc-2012-test}, we show the top methods on the public
leaderboard as of the time of submission.  Our detector obtains a mAP of 76.4\%,
which is several points higher than the next best submission, and is the most
accurate for most categories.

\subsection{MS COCO}
\label{sec:results-mscoco}

Microsoft has recently released the Common Objects in Context dataset, which
contains 80k training images (``2014 train'') and 40k validation images (``2014
val'').  There is an associated MS COCO challenge with a new evaluation metric,
that averages mAP over different IoU thresholds, from 0.5 to 0.95 (written as
``0.5:0.95'').  This places a significantly larger emphasis on localization
compared to the PASCAL VOC metric which only requires IoU of 0.5.

We are only aware of one baseline performance number for this dataset, as
published in the Fast R-CNN paper, which cites a mAP of 19.7\% on the
2015 test-dev set~\cite{fast-rcnn}.  We trained our own Fast R-CNN model on
``2014 train'' using our longer training schedule and obtained a higher mAP of
20.5\% mAP on the same set, which we use as a baseline.
As shown in Table~\ref{tab:coco-2015-test}, when trained on the same images with
the same schedule, our method obtains a large improvement over the baseline with
a mAP of \BestCOCOTest{}\%.

We tried applying the same bounding box voting scheme~\cite{MR-CNN} to COCO, but
found that performance \textit{decreases} on the COCO metric (IOU 0.5:0.95,
second row of Table~\ref{tab:coco-2015-test}).  Interestingly, the scheme increases
performance at IoU 0.5 (the PASCAL metric).  Since the scheme heuristically
blurs together box locations, it can find the general location of objects, but
cannot predict precise box locations, which is important for the new COCO metric.
\newtext{%
  As described in the Appendix, we fixed this for our competition submission by
  raising the voting IoU threshold from 0.5 to $\sim 0.85$.
}

\newtext{%
  We submitted ION to the 2015 MS COCO Detection Challenge and won the Best
  Student Entry with $3^\text{rd}$ place overall.  Using only a single model
  (no ensembling), our submission achieved 31.0\% on test-competition
  score and 31.2\% on test-dev score (Table~\ref{tab:coco-2015-test}).
  After the competition, we further improved our test-dev score to 33.1\% by
  adding left-right flipping and adjusting training parameters.
  See the Appendix for details on our challenge submission.
  %To get from 24.9\% to 32.3\%, we used a mix
  %of MCG\cite{MCG-PABMM2015} and RPN\cite{ren2015faster} proposals, two extra
  %convolution layers, longer training with higher momentum, left-right flipping
  %at test time, and improved NMS and weighted voting thresholds.  See the
  %Appendix for details.
}

\subsection{Improvement for small objects}

In general, small objects are challenging for detectors: there are fewer pixels
on the object, they are harder to localize, and there can be many more of them
per image.  Small objects are even more challenging for proposal methods.  For
all experiments, we are using selective search~\cite{UijlingsIJCV2013} for
object proposals, which performs very poorly on small objects in COCO with an
average recall under 10\%~\cite{deepmask}.

We find that our detector shows a large relative improvement in this category.
For COCO, if we look at small\footnote{``Small'' means area $\le 32^2$
px; about 40\% of COCO is ``small.''} objects, average precision and average recall
improve from 4.1\% to \newtext{7.0\%} and from 7.3\% to \newtext{10.7\%} respectively.  We highlight
that this is even higher than the baseline proposal method, which is only
possible because we perform bounding box regression to predict improved box
locations.  Similarly, we show a size breakdown for VOC2007 test in
Figure~\ref{fig:xs-objects} using Hoiem's toolkit for diagnosing
errors~\cite{hoiem12error}, and see similarly large improvements on this dataset as well.
%\kb{Great figure. Can we put a number of relative improvement over the 3 size
%categories? 65\% relative improvement for small when it goes from 4.1 to 6.8,
%etc.? }

\section{Design evaluation}

%%%%%%%%%%%%%%%%%%%%%%%%%%%%%%%%%%%%%%%%%%%%%%
% Move as needed.
\begin{table}[t]
  \centering
  \small{%
    \begin{tabular}{%
        c@{\hskip 0.7em}c@{\hskip 0.7em}c@{\hskip 0.7em}c|rc
      }
      \toprule
      \multicolumn{4}{c|}{\textbf{ROI pooling from:}} &
      \multicolumn{2}{c}{\textbf{Merge features using:}} \\
      C2 & C3 & C4 & C5 &
      {1x1} & {L2+Scale+1x1} \\
      \midrule
      & & & \checkmark & *70.8 & 71.5\\
      & & \checkmark & \checkmark & 69.7 & 74.4 \\
      & \checkmark & \checkmark & \checkmark & 63.6 & \textbf{74.6} \\
      \checkmark & \checkmark & \checkmark & \checkmark & 59.3 & \textbf{74.6} \\
      \bottomrule
    \end{tabular}
  }
  \vspace{6pt}
  \caption{%
    \textbf{Combining features from different layers.}
    Metric: Detection mAP on VOC07 test.
    Training set: 07 trainval + 12 trainval.
    \textbf{1x1}: combine features from different layers using a 1x1
    convolution.
    \textbf{L2+Scale+1x1}: use L2 normalization, scaling (initialized to 1000),
    and 1x1 convolution, as described in section~\ref{sec:pool-multiple}.
    These results do not include ``context features.''
    *This entry is the same as Fast R-CNN~\cite{fast-rcnn}, but trained with our
    hyperparameters.
  }
  \label{tab:pool-multiple}
\end{table}

%%%%%%%%%%%%%%%%%%%%%%%%%%%%%%%%%%%%%%%%%%%%%%

In this section, we explore changes to our architecture and justify our design
choices with experiments on PASCAL VOC 2007.  All numbers in this section are
VOC 2007 test mAP, trained on 2007 trainval + 2012 trainval, with the settings
described in Section~\ref{sec:experimental-setup}.  Note that for this section,
we use dropout in all networks, and a single round of bounding box regression at
test time.

\subsection{Pool from which layers?}

As described in Section~\ref{sec:pool-multiple}, our detector pools regions of
interest (ROI) from multiple layers and combines the result.  A straightforward
approach would be to concatenate the ROI from each layer and reduce the
dimensionality using a 1x1 convolution.  As shown in Table
\ref{tab:pool-multiple} (left column), this does not work.
In VGG16, the convolutional features at different layers can have very different
amplitudes, so that naively combining them leads to unstable learning.  While it
is possible in theory to learn a model with inputs of very different
amplitude, this is ill-conditioned and does not work well in practice.
It is necessary to normalize the amplitude such that the features being pooled
from all layers have similar magnitude.  Our method's normalization scheme fixes
this problem, as shown in Table~\ref{tab:pool-multiple} (right column).

%%%%%%%%%%%%%%%%%%%%%%%%%%%%%%%%%%%%%%%%%%%%%%
% Move as needed.
\begin{table}[t]
  \centering
  \small{%
    \begin{tabular}{l|c|cc}
      \toprule
      \multirow{2}{*}{\textbf{L2 Normalization method}} &
      \multirow{2}{*}{\textbf{Seg.}} &
      \multicolumn{2}{c}{\textbf{Scale:}} \\
      & & {Learned} & {Fixed} \\
      \midrule
      Sum across channels & \checkmark & 76.4 & 76.2 \\
      Sum over all entries & \checkmark & {76.5} & \textbf{76.6}\\
      \bottomrule
    \end{tabular}
  }
  \vspace{6pt}
  \caption{%
    \textbf{Approaches to normalizing feature amplitude.}
    Metric: detection mAP on VOC07 test.
    All methods are regularized with loss from predicting segmentation.
    %combine features from conv3, conv4,
    %conv5, and the IRNN, and the IRNN is regularized with segmentation loss.
  }
  \label{tab:norm}
\end{table}

%%%%%%%%%%%%%%%%%%%%%%%%%%%%%%%%%%%%%%%%%%%%%%

\subsection{How should we normalize feature amplitude?}
\label{sec:norm}

When performing L2 normalization, there are a few choices to be made: do you
sum over channels and perform one normalization per spatial location (as in
ParseNet~\cite{parsenet}), or should you sum over all entries in each pooled ROI
and normalize it as a single blob.  Further, when re-scaling the features back
to an fc6-compatible magnitude, should you use a fixed scale or should you learn a
scale per channel?  The reason why you might want to learn a scale per channel
is that you get more sharing than you would if you relied on the 1x1 convolution
to model the scale.  We evaluate this in Table~\ref{tab:norm}, and find that all
of these approaches perform about the same, and the distinction doesn't matter
for this problem.  The important aspect is whether amplitude is taken into
account; the different schemes we explored in Table~\ref{tab:norm} are all
roughly equivalent in performance.
%Since a fixed scale is simpler to implement
%(and a hair better), we recommend using this in the future, though we note that
%all other experiments use a learned scale (bottom left table entry).

To determine the initial scale, we measure the mean scale of features pooled
from conv5 on the training set, and use that as the fixed scale.  Using Fast
R-CNN, we measured the mean norm to be approximately 1000 when summing over all
entries, and 130 when summing across channels.

%%%%%%%%%%%%%%%%%%%%%%%%%%%%%%%%%%%%%%%%%%%%%%
% Move as needed.
\begin{table}[t]
  \centering
  \small{%
    \begin{tabular}{c@{\hskip 0.7em}c@{\hskip 0.7em}c@{\hskip 0.7em}c@{\hskip 0.7em}c|rr}
      \toprule
      \multicolumn{5}{c|}{\textbf{ROI pooling from:}} &
      \multicolumn{2}{c}{\textbf{Use seg. loss?}} \\
      C2 & C3 & C4 & C5 & IRNN & No & Yes \\
      \midrule
      & & & & \checkmark & 69.9 & 70.6 \\
      & & & \checkmark & \checkmark & 73.9 & \newtext{74.2} \\
      & & \checkmark & \checkmark & \checkmark & 75.1 & 76.2 \\
      & \checkmark & \checkmark & \checkmark & \checkmark & 75.6 & {76.5} \\
      \checkmark & \checkmark & \checkmark & \checkmark & \checkmark & 74.9 & \textbf{76.8} \\
      \bottomrule
    \end{tabular}
  }
  \vspace{6pt}
  \caption{%
    \textbf{Effect of segmentation loss.}
    Metric: detection mAP on VOC07 test.  Adding segmentation loss tends to improve
    detection performance by about 1 mAP, with no test-time penalty.
    %*We
%believe that this is an outlier and possible bug, as it is inconsistent with
%other experiments; we will address this in future revisions.
    %\todo{Re-compute 70.2 result, explain outlier if needed.}
%\kb{Larry and I are in favor of just reporting it and neither calling it out
%nor omitting it. Make a call.}
  }
  \label{tab:seg}
\end{table}

%%%%%%%%%%%%%%%%%%%%%%%%%%%%%%%%%%%%%%%%%%%%%%

\subsection{How much does segmentation loss help?}
\label{sec:seg}

Although our target task is object detection, many datasets also have semantic
segmentation labels, where the object class of every pixel is labeled.  Many
images in PASCAL VOC and every image in COCO has these labels.  This is valuable
information that can be incorporated into a training algorithm to improve
performance.

As shown in Figure~\ref{fig:arch-rnn}, when adding stacked IRNNs it is possible
to have them also predict a semantic segmentation output---a multitask setup.
In Table~\ref{tab:seg}, we see that these extra labels consistently provide
about a +1 point boost in mAP for object detection.  This is because we are
training the network with more bits of supervision, so even though we are adding
extra labels that we do not care about during inference, the features inside the
network are trained to contain more information than they would have otherwise
if only trained on object detection.  Since this is an extra layer used only for
training, we can drop the layer at test time and get a +1 mAP point boost with no
change in runtime.

\subsection{How should we incorporate context?}

\begin{figure}[t]
  \begin{center}
   \includegraphics[width=0.99\linewidth]{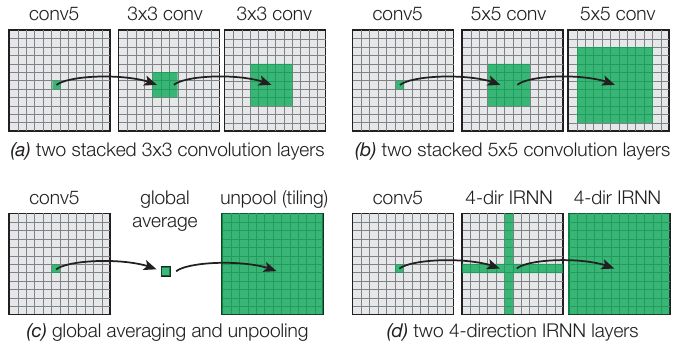}
  \end{center}
  \vspace{-12pt}
  \caption{%
    \textbf{Receptive field of different layer types.}
    When considering a single cell in the input, what output cells depend on it?
    \textbf{(a)} If we add two stacked 3x3 convolutions on top of conv5, then a cell in
    the input influences a 5x5 window in the output.  \textbf{(b)} Similarly, for a 5x5
    convolution, one cell influences a 9x9 window in the output.  \textbf{(c)} For global
    average pooling, every cell in the output depends on the entire input, but
    the output is the same value repeated.  \textbf{(d)} For IRNNs, every cell in the
    output depends on the entire input, but also varies spatially.
  }
  \label{fig:global}
\end{figure}

While RNNs are a powerful mechanism of incorporating context, they are not the only
method.  For example, one could simply add more convolutional layers on top of
conv5 and then pool out of the top convolutional layer.  As shown in
Figure~\ref{fig:global}, stacked 3x3 convolutions add two cells worth of
context, and stacked 5x5 convolutions add 6 cells.  Alternatively, one could
use a global average and unpool (tile or repeat spatially) back to the original
shape as in ParseNet~\cite{parsenet}.
%This is equivalent to tiling it to a 512x7x7 feature and including it in
%the list of features that get combined.

We compared these approaches on VOC 2007 test, shown in Table~\ref{tab:global}.
The 2x stacked 4-dir IRNN layers have fewer parameters than the alternatives, and
perform better on the test set (both with and without segmentation labels).
Therefore, we use this architecture to compute ``context features'' for all
other experiments.

%%%%%%%%%%%%%%%%%%%%%%%%%%%%%%%%%%%%%%%%%%%%%%
% Move as needed.

%%%%%%%%%%%%%%%%%%%%%%%%%%%%%%%%%%%%%%%%%%%%%%

%%%%%%%%%%%%%%%%%%%%%%%%%%%%%%%%%%%%%%%%%%%%%%
% Move as needed.
\begin{table}[t]
  \centering
  \small{%
    \begin{tabular}{lcl}
      \toprule
      \textbf{Context method} & \textbf{Seg.} & \textbf{mAP} \\
      \midrule
      (a) 2x stacked 512x3x3 conv & & 74.8 \\
      (b) 2x stacked 256x5x5 conv & & 74.6 \\
      (c) Global average pooling & & 74.9 \\
      (d) 2x stacked 4-dir IRNN & & \textbf{75.6} \\
      \midrule
      (a) 2x stacked 512x3x3 conv & \checkmark & 75.2 \\
      (d) 2x stacked 4-dir IRNN & \checkmark & \textbf{76.5} \\
      \bottomrule
    \end{tabular}
  }
  \vspace{6pt}
  \caption{%
    \textbf{Comparing approaches to adding context.}
    All rows also pool out of conv3, conv4, and conv5.
    Metric: detection mAP on VOC07 test.
    \textbf{Seg:} if checked, the top layer received extra supervision from
    semantic segmentation labels.
  }
  \label{tab:global}
  \vspace{-0.1em}
\end{table}

%%%%%%%%%%%%%%%%%%%%%%%%%%%%%%%%%%%%%%%%%%%%%%

\subsection{Which IRNN architecture?}
\label{sec:irnn-arch}

When designing the IRNN for incorporating context, there are a few basic decisions
to be made, namely how many layers and how many hidden units per layer.
In addition, we explore the idea of entirely removing the recurrent transition
(equivalent to replacing it with the identity matrix), so that
the IRNN consist of repeated steps of: accumulate, ReLU, accumulate, \etc.
Note that this is not the same as an integral/area image, since each step has
ReLU.

As shown in Table~\ref{tab:num-irnn-layers}, using 2 IRNN layers performs the
best on VOC 2007 test.  While stacking more convolution layers tends to make
ConvNets perform better, the same is not always true for
RNNs~\cite{karpathy15rnn}.
We also found that the number of hidden units did not have a strong effect on
the performance (Table~\ref{tab:num-hidden-units}), and chose 512 as the
baseline size for all other experiments.

Finally, we were surprised to discover that removing the recurrent
$\mathbf{W}_{hh}$ transition performs almost as well as learning it
(Table~\ref{tab:num-hidden-units}).  It seems that the input-to-hidden and
hidden-to-output connections contain sufficient context that the
recurrent transition can be removed and replaced with an addition,
saving a large matrix multiply.

%%%%%%%%%%%%%%%%%%%%%%%%%%%%%%%%%%%%%%%%%%%%%%
% Move as needed.
\begin{table}[t]
  \centering
  \small{%
    \begin{tabular}{%
        c@{\hskip 0.7em}c@{\hskip 0.7em}c@{\hskip 0.7em}c@{\hskip 0.7em}c
        |c|ccc
      }
      \toprule
      \multicolumn{5}{c|}{\textbf{ROI pooling from:}} &
      \multirow{2}{*}{\textbf{Seg.}} &
      \multicolumn{3}{c}{\textbf{\# IRNN layers}}
      \\
      C2 & C3 & C4 & C5 & IRNN & & 1 & 2 & 3 \\
      \midrule
      & & & & \checkmark & \checkmark & & 70.6 & \\
      & & & \checkmark & \checkmark & \checkmark & 74.3 & & \\
      & & \checkmark & \checkmark & \checkmark & \checkmark & 75.8 & 76.2 & \\
      & \checkmark & \checkmark & \checkmark & \checkmark & \checkmark & 76.1 & {76.5} & 75.9 \\
      \checkmark & \checkmark & \checkmark & \checkmark & \checkmark &
      \checkmark & & \textbf{76.8} &\\
      \bottomrule
    \end{tabular}
  }
  \vspace{6pt}
  \caption{%
    \textbf{Varying the number of IRNN layers.}
    Metric: mAP on VOC07 test.
    Segmentation loss is used to regularize the top IRNN layer.
    All IRNNs use 512 hidden units.
  }
  \label{tab:num-irnn-layers}
\end{table}

%%%%%%%%%%%%%%%%%%%%%%%%%%%%%%%%%%%%%%%%%%%%%%

%%%%%%%%%%%%%%%%%%%%%%%%%%%%%%%%%%%%%%%%%%%%%%
% Move as needed.
\begin{table}[t]
  \centering
  \small{%
    \begin{tabular}{
      c@{\hskip 0.7em}c@{\hskip 0.7em}c@{\hskip 0.7em}c
      |c|c|cc}
      \toprule
      \multicolumn{4}{c|}{\textbf{ROI pooling from:}} &
      \multirow{2}{*}{\textbf{Seg.}} &
      \multirow{2}{*}{\textbf{\# units}} &
      \multicolumn{2}{c}{\textbf{Include $\mathbf{W}_{hh}$?}} \\
      C3 & C4 & C5 & IRNN & & & Yes & No \\
      \midrule
      \checkmark & \checkmark & \checkmark & \checkmark & \checkmark &128
      & 76.4 &75.5 \\
      \checkmark & \checkmark & \checkmark & \checkmark & \checkmark &256
      & \textbf{76.5}  & 75.3 \\
      \checkmark & \checkmark & \checkmark & \checkmark & \checkmark &512
      & \textbf{76.5} & 76.1 \\
      \checkmark & \checkmark & \checkmark & \checkmark & \checkmark &1024
      & 76.2 & {76.4} \\
      \bottomrule
    \end{tabular}
  }
  \vspace{6pt}
  \caption{%
    \textbf{Varying the hidden transition.}
    We vary the number of units and try either learning recurrent transition
    $\mathbf{W}_{hh}$ initialized to the identity, or entirely removing it
    (same as setting $\mathbf{W}_{hh} = I$).
  }
  \label{tab:num-hidden-units}
\end{table}

%%%%%%%%%%%%%%%%%%%%%%%%%%%%%%%%%%%%%%%%%%%%%%

\subsection{Other variations}

There are some other variations on our architecture that perform almost as
well, which we summarize in Table~\ref{tab:other}.  For example, (a) the first
IRNN only processes two directions left/right and the second IRNN only processes
up/down.  This kind of operation was explored in ReNet~\cite{renet} and
performs the same as modeling all four directions in both IRNN layers.  We also
explored (b) pooling out of both IRNNs, and (c) pooling out of both stacked
convolutions and the IRNNs.  None of these variations perform better than our
main method.

%%%%%%%%%%%%%%%%%%%%%%%%%%%%%%%%%%%%%%%%%%%%%%
% Move as needed.
\begin{table}[t]
  \centering
  \small{%
    \begin{tabular}{llr}
      \toprule
      \textbf{Variation} & \textbf{mAP} \\
      \midrule
      Our method & \textbf{76.5} \\
      (a) Left-right then up-down & \textbf{76.5} \\
      (b) Pool out of both IRNNs & 75.9 \\
      (c) Combine 2x stacked 512x3x3 conv and IRNN & \textbf{76.5} \\
      \bottomrule
    \end{tabular}
  }
  \vspace{6pt}
  \caption{%
    \textbf{Other variations.}  Metric: VOC07 test mAP.  We list some other
    variations that all perform about the same.
  }
  \label{tab:other}
\end{table}

%%%%%%%%%%%%%%%%%%%%%%%%%%%%%%%%%%%%%%%%%%%%%%

\section{Conclusion}
This paper introduces the Inside-Outside Net (ION), an architecture that leverages
context and multi-scale knowledge for object detection. Our architecture uses a
2x stacked 4-directional IRNN for context, and multi-layer ROI pooling with
normalization for improved object description.  To justify our design choices,
we conducted extensive experiments evaluating choices like the number of layers
combined, using segmentation loss, normalizing feature amplitudes, different
IRNN architectures, and other variations.  We achieve state-of-the-art results
on both PASCAL VOC and COCO, and find our proposed architecture is particularly
effective at improving detection of small objects.

\section*{Acknowledgements}
This project was the result of an internship at Microsoft Research (MSR).  We
would like to thank Abhinav Shrivastava and Ishan Misra for helpful discussions
while at MSR.  We thank NVIDIA for the donation of K40 GPUs.

% comment out so it builds faster
%\input{example_results}

%\input{lorem}

{\small
\bibliographystyle{ieee}
\bibliography{submission}
}

\vfill
\newpage
\section*{Appendix: 2015 MS COCO Competition}

\begin{table*}[t]
  \centering
  \small{\begin{tabular}{%
  lrrrrrr}
  \toprule
  \multirow{2}{*}{\textbf{Method}} &
  \multirow{2}{*}{\textbf{Proposals}} &
  \textbf{Train:} &
  train & train &
  trainval35k &
  trainval35k \\
  & &
  \textbf{Test:} &
  minival &
  test-dev &
  minival &
  test-dev \\
  \midrule
  baseline VGG16 & Sel. Search & & 20.3 & 20.5 & & \\
  ION VGG16 & Sel. Search & & 24.4 & 24.9 & & \\
  ION VGG16 + conv7& Sel. Search & & & 25.1 & & \\
  ION VGG16 + conv7& MCG + RPN & & & & 28.4 & 29.0 \\
  ~+ \textbf{W} (box refinement) & & & & & 30.0 & 30.6 \\
  ~+ flipping + more, better training & & & & & 32.5 & \textbf{33.1} \\
  \bottomrule
\end{tabular}
  \vspace{3pt}
  \caption{%
    Breakdown of gains for the post-competition model. The reported metric is
    Avg.\@ Precision, IoU: 0.5:0.95. The training set ``trainval35k'' includes all
    of train together with approximately 35k images from val, after removing the
    5k minival set.  All entries use a single ConvNet model (no ensembling).
    The majority of the gains come from the ION model (20.5 $\rightarrow$ 24.9)
    and better proposals with more training data (MCG + RPN: 25.1 $\rightarrow$ 29.0).
    Two rounds of bounding box regression with weighted voting and longer
    training with improved hyperparameters also yield important gains.  Note
    that we use a modified version of RPN~\cite{ren2015faster}, described in the
    Appendix text.
  }
  \label{tab:coco-breakdown}
}

\end{table*}

In this section, we describe our submission to the 2015 MS COCO Detection
Challenge, which won Best Student Entry and finished $3^\text{rd}$ place
overall, with a score of 31.0\% mAP on 2015 test-challenge and 31.2\% on
2015 test-dev.  Later in this section we describe further post-competition
improvements to achieve 33.1\% on test-dev.  Both models use a single ConvNet
(no ensembling).

For our challenge submission, we made several improvements:
used a mix of MCG (Multiscale Combinatorial Grouping~\cite{MCG-PABMM2015}) and
RPN (Region Proposal Net~\cite{ren2015faster}) proposal boxes, added two extra
512x3x3 convolutional layers, trained for longer, and used two
rounds of bounding box regression with a modified version of weighted
voting~\cite{MR-CNN}.  At test time, our model runs in 2.7 seconds/image on a
single Titan X GPU (excluding proposal generation).

We describe all changes in more detail below (note that many of these choices
were driven by the need to meet the challenge deadline, and thus may be
suboptimal):
\begin{packed_enum}
  \item \textbf{Train+val.}  For the competition, we train on both train and
    validation sets.  We hold out 5000 images from the validation as our new
    validation set called ``minival.''
  \item \textbf{MCG+RPN box proposals.} We get the largest improvement by
    replacing selective search with a mix of MCG~\cite{MCG-PABMM2015} and
    RPN~\cite{ren2015faster} boxes.  We modify RPN from the baseline
    configuration described in
    \cite{ren2015faster} by adding more anchor boxes, in particular
    smaller ones, and using a mixture of 3x3 (384) and 5x5 (128) convolutions.
    Our anchor configuration uses a total of 22 anchors per location
    with the following shapes:
    32x32 and aspect ratios \{1:2, 1:1, 2:1\} $\times$ scales \{64, 90.5, 128,
    181, 256, 362, 512\}.
    We also experiment with predicting proposals from concatenated conv4\_3
    and conv5\_3 features (after L2 normalization and scaling), rather than
    from conv5\_3 only. We refer to these two configurations as RPN1
    (without concatenated features) and RPN2 (with concatenated features).
%    \\
%    RPN1: rpn\_vgg16\_384\_3x3\_128\_5x5\_all\_anchors
%    \\
%    RPN2: rpn\_vgg16\_384\_3x3\_128\_5x5\_all\_anchors\_cat45
%    \\
    Using VGG16, RPN1 and RPN2 achieve average recalls of 44.1\% and 44.3\%, respectively,
    compared to selective search, 41.7\%, and MCG, 51.6\%.
    Despite their lower average recalls, we found that RPN1 and RPN2 give
    comparable detection results to MCG.
    We also found that mixing 1000 MCG boxes with 1000 RPN1 or RPN2 boxes
    performs even better than 2000 of either method alone, suggesting that the
    two methods are complementary.
    Thus, we train with 1000 RPN1 and 1000 MCG boxes. At test time, we use 1000
    RPN2 and 2000 MCG boxes, which gives a +0.3 mAP improvement on minival
    compared to using the same boxes as training.
  \item \textbf{conv6+conv7.}  We use the model listed in row (c) of
    Table~\ref{tab:other}: two 512x3x3 convolutions on top of conv5\_3 which we
    call ``conv6'' and ``conv7''.  We also pool out of conv7, so in total we
    pool out of conv3\_3, conv4\_3, conv5\_3, conv7, and IRNN.
  \item \textbf{Longer training.}  Since training on COCO is very slow, we
  explored ideas by initializing new networks from the weights of the previous
  best network.  While we have not tried training our best configuration
  starting from VGG16 ImageNet weights, we do not see any reason why it would not
  work.  Here is the exact training procedure we used:
  \begin{packed_enum}
    \item Train using 2000 selective search boxes, using the schedule described
      in Section~\ref{sec:experimental-setup}, but stopping after 220k iterations
      of fine-tuning.  This achieves 24.8\% on 2015 test-dev.
    \item Change the box proposals to a mix of 1000 MCG boxes and 1000 RPN1
      boxes.  Train for 100k iterations with conv layers frozen (learning rate: exp decay
      $5 \cdot 10^{-3} \rightarrow 10^{-4}$), and for 80k iterations with
      only conv1 and conv2 frozen (learning rate: exp decay $10^{-3} \rightarrow 10^{-5}$).
  \end{packed_enum}
  \item \textbf{2xBBReg + WV.}  We use the iterative bounding box regression and
    weighted voting scheme described in \cite{MR-CNN}, but as noted in
    Section~\ref{sec:results-mscoco}, this does not work out-of-the-box for COCO
    since boxes are blurred together and precise localization is lost.  We solve
    this by adjusting the thresholds so that only very similar boxes are blurred
    together with weighted voting.  We jointly optimized the NMS (non-max
    suppression) and voting threshold on the validation set, by evaluating all
    boxes once and then randomly sampling hundreds of thresholds.  On our
    minival set, the optimal IoU (intersection-over-union) threshold is 0.443
    for NMS and 0.854 for weighted voting.  Compared to using a single round of
    standard NMS (IoU threshold 0.3), these settings give +1.3 mAP on minival.
%
  %\item \textbf{Deduplication.}  Finally, Fast R-CNN de-duplicates boxes that
    %are within 16 pixels of each other, since they map to the same grid cells on
    %conv5\_3.  Since ION pools from higher-resolution layers, we change the
    %de-duplication rule to only merge boxes within 8 pixels of each other.
    %However, we later found that this has nearly zero effect on accuracy, so it
    %the default 16-pixel rule could be used in the future.
%
\end{packed_enum}

\subsection*{Post-competition improvements}
We have an improved model which did not finish in time for the challenge, which
achieves 33.1\% on 2015 test-dev.  We made these further adjustments:
\begin{packed_enum}
  \item \textbf{Training longer with 0.99 momentum.}  We found that if the
    momentum vector is reset when re-starting training, the accuracy drops by
    several points and takes around 50k iterations to recover.  Based on this
    observation, we increased momentum to 0.99 and correspondingly decreased
    learning rate by a factor of 10.  We initialized from our competition
    submission (above), and:
    \begin{packed_enum}
      \item further trained for
      80k iterations with only conv1 and conv2 frozen, (learning rate: exp decay
      $10^{-4} \rightarrow 10^{-6}$).
      By itself, this gives a boost of +0.4 mAP on minival.
    \item We trained for another 160k iterations with no layers frozen (learning
      rate: exp decay $10^{-4} \rightarrow 10^{-6}$) which gives +0.7 mAP on
      minival.
  \end{packed_enum}
  \item \textbf{Left-right flipping.}  We evaluate the model twice, once on the
    original image, and once on the left-right flipped image.  To merge the
    predictions, we average both the softmax scores and box regression shifts
    (after flipping back).  By itself, this gives a boost of +0.8 mAP on minival.
  \item \textbf{More box proposals.} At test time, we use 4000 proposal boxes
    (2000 RPN2 and 2000 MCG boxes).  For the model submitted to the competition,
    this performs +0.1 mAP better than 3000 boxes (1000 RPN2 and 2000 MCG
    boxes).
\end{packed_enum}
At test time, the above model runs in 5.5 seconds/image on a single Titan X GPU
(excluding proposal generation).  Most of the slowdown is from the left-right
flipping. Table~\ref{tab:coco-breakdown} provides a breakdown of the gains due
to the various competition and post-competition changes.

% put here for now so we can visualize it
%\input{supplemental}

\end{document}

%\title{Detecting Objects in Context with Multi-layer ROI Pooling and Recurrent
%Neural Networks.}
%\title{Detecting Objects in Context with the Inside-Outside Net}
\title{%
  \vspace{-0.43em}
  Supplemental Material:
  Inside-Outside Net:
  Detecting Objects in Context with Skip Pooling and
  Recurrent Neural Networks
  \vspace{-0.43em}
}
%Skip Pooling and Recurrent Neural Networks}

%% omit the following lines up until the closing ``}''.
%% Additional authors and addresses can be added with ``\and'',
%% just like the second author.
%% To save space, use either the email address or home page, not both
%\and

\maketitle

\section*{Visual results on COCO test-dev}

In the directory \texttt{coco-test-dev-detections/}, we show example detections
on COCO 2015 test-dev (first 100 images), using our ION model trained on COCO
2014 train (validation images were not used).  Boxes are annotated with their
softmax score.  All boxes within 50\% of the max score for that image are shown,
so low-scoring boxes may be shown for some images.

\section*{Visual results on VOC 2007 test}

Similarly, \texttt{voc-2007-test-detections/} shows example detections on VOC
2007 test (first 100 images), using our ION model trained on 2007 trainval +
2012 trainval.  Boxes are annotated with their softmax score.  All boxes within
50\% of the max score for that image are shown, so low-scoring boxes may be
shown for some images.

\section*{Additional plots for VOC 2007 error diagnosis}

In this document, we include the full error analysis for ION (our object
detector), tested on PASCAL VOC 2007 test, trained on 2007 trainval + 2012 trainval.
These plots are from Hoiem's toolkit for diagnosing errors~\cite{hoiem12error}.
Figure~{5} in the main paper is a condensed version of these plots, focusing
only on area.

From these plots, we can see that we see large gains across the board, but
particularly for small objects.

% comment out so it builds faster
%\input{example_results}

%\input{lorem}

{\small
\bibliographystyle{ieee}
\bibliography{submission}
}

%\vspace{16pt}

\begin{figure}[t]
  \begin{center}
   \includegraphics[height=0.35\linewidth]{figures/hoiem/noseg_conv5_thresh0_sched2_0712tv/plots_impact_strong.pdf}
   \includegraphics[height=0.35\linewidth]{figures/hoiem/noseg_conv5normconcat_thresh0_sched2_0712tv/plots_impact_strong.pdf}
   \includegraphics[height=0.35\linewidth]{figures/hoiem/noseg_conv45normconcat_thresh0_sched2_0712tv/plots_impact_strong.pdf}
   \includegraphics[height=0.35\linewidth]{figures/hoiem/noseg_conv345normconcat_thresh0_sched2_0712tv/plots_impact_strong.pdf}
   \includegraphics[height=0.35\linewidth]{figures/hoiem/rnn_layer2_x3rcnn_conv345normconcat_thresh0_sched2_in_0712tv/plots_impact_strong.pdf}
   \includegraphics[height=0.35\linewidth]{figures/hoiem/rnn_layer2_x3rcnn_conv345normconcat_thresh0_sched2_bbreg2wvall_bbthresh01abs_in_0712tv/plots_impact_strong.pdf}
   \includegraphics[height=0.35\linewidth]{figures/hoiem/rnn_layer2_x3rcnn_conv345normconcat_thresh0_nodrop_sched2_bbreg2wvall_bbthresh01abs_in_0712tv/plots_impact_strong.pdf}
  \end{center}
  \vspace{-12pt}
  \caption{%
    \textbf{Impact.}  We can see that as our model complexity increases, the impact due
    to size decreases.
  }
  \label{fig:hoiem}
\end{figure}

\begin{figure}[t]
  \begin{center}
   \includegraphics[height=0.35\linewidth]{figures/hoiem/noseg_conv5_thresh0_sched2_0712tv/plots_area_strong.pdf}
   \includegraphics[height=0.35\linewidth]{figures/hoiem/noseg_conv5normconcat_thresh0_sched2_0712tv/plots_area_strong.pdf}
   \includegraphics[height=0.35\linewidth]{figures/hoiem/noseg_conv45normconcat_thresh0_sched2_0712tv/plots_area_strong.pdf}
   \includegraphics[height=0.35\linewidth]{figures/hoiem/noseg_conv345normconcat_thresh0_sched2_0712tv/plots_area_strong.pdf}
   \includegraphics[height=0.35\linewidth]{figures/hoiem/rnn_layer2_x3rcnn_conv345normconcat_thresh0_sched2_in_0712tv/plots_area_strong.pdf}
   \includegraphics[height=0.35\linewidth]{figures/hoiem/rnn_layer2_x3rcnn_conv345normconcat_thresh0_sched2_bbreg2wvall_bbthresh01abs_in_0712tv/plots_area_strong.pdf}
   \includegraphics[height=0.35\linewidth]{figures/hoiem/rnn_layer2_x3rcnn_conv345normconcat_thresh0_nodrop_sched2_bbreg2wvall_bbthresh01abs_in_0712tv/plots_area_strong.pdf}
  \end{center}
  \vspace{-12pt}
  \caption{%
    \textbf{VOC 2007 normalized AP by area.}
  }
  \label{fig:hoiem}
\end{figure}

\begin{figure}[t]
  \begin{center}
   \includegraphics[height=0.35\linewidth]{figures/hoiem/noseg_conv5_thresh0_sched2_0712tv/plots_truncation_strong.pdf}
   \includegraphics[height=0.35\linewidth]{figures/hoiem/noseg_conv5normconcat_thresh0_sched2_0712tv/plots_truncation_strong.pdf}
   \includegraphics[height=0.35\linewidth]{figures/hoiem/noseg_conv45normconcat_thresh0_sched2_0712tv/plots_truncation_strong.pdf}
   \includegraphics[height=0.35\linewidth]{figures/hoiem/noseg_conv345normconcat_thresh0_sched2_0712tv/plots_truncation_strong.pdf}
   \includegraphics[height=0.35\linewidth]{figures/hoiem/rnn_layer2_x3rcnn_conv345normconcat_thresh0_sched2_in_0712tv/plots_truncation_strong.pdf}
   \includegraphics[height=0.35\linewidth]{figures/hoiem/rnn_layer2_x3rcnn_conv345normconcat_thresh0_sched2_bbreg2wvall_bbthresh01abs_in_0712tv/plots_truncation_strong.pdf}
   \includegraphics[height=0.35\linewidth]{figures/hoiem/rnn_layer2_x3rcnn_conv345normconcat_thresh0_nodrop_sched2_bbreg2wvall_bbthresh01abs_in_0712tv/plots_truncation_strong.pdf}
  \end{center}
  \vspace{-12pt}
  \caption{%
    \textbf{VOC 2007 normalized AP by truncation.}  N: not truncated, T: truncated.
  }
  \label{fig:hoiem}
\end{figure}

\begin{figure}[t]
  \begin{center}
   \includegraphics[height=0.35\linewidth]{figures/hoiem/noseg_conv5_thresh0_sched2_0712tv/plots_aspect_strong.pdf}
   \includegraphics[height=0.35\linewidth]{figures/hoiem/noseg_conv5normconcat_thresh0_sched2_0712tv/plots_aspect_strong.pdf}
   \includegraphics[height=0.35\linewidth]{figures/hoiem/noseg_conv45normconcat_thresh0_sched2_0712tv/plots_aspect_strong.pdf}
   \includegraphics[height=0.35\linewidth]{figures/hoiem/noseg_conv345normconcat_thresh0_sched2_0712tv/plots_aspect_strong.pdf}
   \includegraphics[height=0.35\linewidth]{figures/hoiem/rnn_layer2_x3rcnn_conv345normconcat_thresh0_sched2_in_0712tv/plots_aspect_strong.pdf}
   \includegraphics[height=0.35\linewidth]{figures/hoiem/rnn_layer2_x3rcnn_conv345normconcat_thresh0_sched2_bbreg2wvall_bbthresh01abs_in_0712tv/plots_aspect_strong.pdf}
   \includegraphics[height=0.35\linewidth]{figures/hoiem/rnn_layer2_x3rcnn_conv345normconcat_thresh0_nodrop_sched2_bbreg2wvall_bbthresh01abs_in_0712tv/plots_aspect_strong.pdf}
  \end{center}
  \vspace{-12pt}
  \caption{%
    aspect
  }
  \label{fig:hoiem}
\end{figure}

\begin{figure}[t]
  \begin{center}
   \includegraphics[height=0.35\linewidth]{figures/hoiem/noseg_conv5_thresh0_sched2_0712tv/plots_height_strong.pdf}
   \includegraphics[height=0.35\linewidth]{figures/hoiem/noseg_conv5normconcat_thresh0_sched2_0712tv/plots_height_strong.pdf}
   \includegraphics[height=0.35\linewidth]{figures/hoiem/noseg_conv45normconcat_thresh0_sched2_0712tv/plots_height_strong.pdf}
   \includegraphics[height=0.35\linewidth]{figures/hoiem/noseg_conv345normconcat_thresh0_sched2_0712tv/plots_height_strong.pdf}
   \includegraphics[height=0.35\linewidth]{figures/hoiem/rnn_layer2_x3rcnn_conv345normconcat_thresh0_sched2_in_0712tv/plots_height_strong.pdf}
   \includegraphics[height=0.35\linewidth]{figures/hoiem/rnn_layer2_x3rcnn_conv345normconcat_thresh0_sched2_bbreg2wvall_bbthresh01abs_in_0712tv/plots_height_strong.pdf}
   \includegraphics[height=0.35\linewidth]{figures/hoiem/rnn_layer2_x3rcnn_conv345normconcat_thresh0_nodrop_sched2_bbreg2wvall_bbthresh01abs_in_0712tv/plots_height_strong.pdf}
  \end{center}
  \vspace{-12pt}
  \caption{%
    \textbf{VOC 2007 normalized AP by height.}
  }
  \label{fig:hoiem}
\end{figure}

\begin{figure}[t]
  \begin{center}
   \includegraphics[height=0.35\linewidth]{figures/hoiem/noseg_conv5_thresh0_sched2_0712tv/plots_person_strong.pdf}
   \includegraphics[height=0.35\linewidth]{figures/hoiem/noseg_conv5normconcat_thresh0_sched2_0712tv/plots_person_strong.pdf}
   \includegraphics[height=0.35\linewidth]{figures/hoiem/noseg_conv45normconcat_thresh0_sched2_0712tv/plots_person_strong.pdf}
   \includegraphics[height=0.35\linewidth]{figures/hoiem/noseg_conv345normconcat_thresh0_sched2_0712tv/plots_person_strong.pdf}
   \includegraphics[height=0.35\linewidth]{figures/hoiem/rnn_layer2_x3rcnn_conv345normconcat_thresh0_sched2_in_0712tv/plots_person_strong.pdf}
   \includegraphics[height=0.35\linewidth]{figures/hoiem/rnn_layer2_x3rcnn_conv345normconcat_thresh0_sched2_bbreg2wvall_bbthresh01abs_in_0712tv/plots_person_strong.pdf}
   \includegraphics[height=0.35\linewidth]{figures/hoiem/rnn_layer2_x3rcnn_conv345normconcat_thresh0_nodrop_sched2_bbreg2wvall_bbthresh01abs_in_0712tv/plots_person_strong.pdf}
  \end{center}
  \vspace{-12pt}
  \caption{%
    \textbf{VOC 2007 normalized AP for ``person.''}
  }
  \label{fig:hoiem}
\end{figure}

\begin{figure}[t]
  \begin{center}
   \includegraphics[height=0.35\linewidth]{figures/hoiem/noseg_conv5_thresh0_sched2_0712tv/plots_pottedplant_strong.pdf}
   \includegraphics[height=0.35\linewidth]{figures/hoiem/noseg_conv5normconcat_thresh0_sched2_0712tv/plots_pottedplant_strong.pdf}
   \includegraphics[height=0.35\linewidth]{figures/hoiem/noseg_conv45normconcat_thresh0_sched2_0712tv/plots_pottedplant_strong.pdf}
   \includegraphics[height=0.35\linewidth]{figures/hoiem/noseg_conv345normconcat_thresh0_sched2_0712tv/plots_pottedplant_strong.pdf}
   \includegraphics[height=0.35\linewidth]{figures/hoiem/rnn_layer2_x3rcnn_conv345normconcat_thresh0_sched2_in_0712tv/plots_pottedplant_strong.pdf}
   \includegraphics[height=0.35\linewidth]{figures/hoiem/rnn_layer2_x3rcnn_conv345normconcat_thresh0_sched2_bbreg2wvall_bbthresh01abs_in_0712tv/plots_pottedplant_strong.pdf}
   \includegraphics[height=0.35\linewidth]{figures/hoiem/rnn_layer2_x3rcnn_conv345normconcat_thresh0_nodrop_sched2_bbreg2wvall_bbthresh01abs_in_0712tv/plots_pottedplant_strong.pdf}
  \end{center}
  \vspace{-12pt}
  \caption{%
    pottedplant
  }
  \label{fig:hoiem}
\end{figure}

\begin{figure}[t]
  \begin{center}
   \includegraphics[height=0.35\linewidth]{figures/hoiem/noseg_conv5_thresh0_sched2_0712tv/plots_chair_strong.pdf}
   \includegraphics[height=0.35\linewidth]{figures/hoiem/noseg_conv5normconcat_thresh0_sched2_0712tv/plots_chair_strong.pdf}
   \includegraphics[height=0.35\linewidth]{figures/hoiem/noseg_conv45normconcat_thresh0_sched2_0712tv/plots_chair_strong.pdf}
   \includegraphics[height=0.35\linewidth]{figures/hoiem/noseg_conv345normconcat_thresh0_sched2_0712tv/plots_chair_strong.pdf}
   \includegraphics[height=0.35\linewidth]{figures/hoiem/rnn_layer2_x3rcnn_conv345normconcat_thresh0_sched2_in_0712tv/plots_chair_strong.pdf}
   \includegraphics[height=0.35\linewidth]{figures/hoiem/rnn_layer2_x3rcnn_conv345normconcat_thresh0_sched2_bbreg2wvall_bbthresh01abs_in_0712tv/plots_chair_strong.pdf}
   \includegraphics[height=0.35\linewidth]{figures/hoiem/rnn_layer2_x3rcnn_conv345normconcat_thresh0_nodrop_sched2_bbreg2wvall_bbthresh01abs_in_0712tv/plots_chair_strong.pdf}
  \end{center}
  \vspace{-12pt}
  \caption{%
    \textbf{VOC 2007 normalized AP for ``chair.''}
  }
  \label{fig:hoiem}
\end{figure}

\begin{figure}[t]
  \begin{center}
   \includegraphics[height=0.35\linewidth]{figures/hoiem/noseg_conv5_thresh0_sched2_0712tv/plots_diningtable_strong.pdf}
   \includegraphics[height=0.35\linewidth]{figures/hoiem/noseg_conv5normconcat_thresh0_sched2_0712tv/plots_diningtable_strong.pdf}
   \includegraphics[height=0.35\linewidth]{figures/hoiem/noseg_conv45normconcat_thresh0_sched2_0712tv/plots_diningtable_strong.pdf}
   \includegraphics[height=0.35\linewidth]{figures/hoiem/noseg_conv345normconcat_thresh0_sched2_0712tv/plots_diningtable_strong.pdf}
   \includegraphics[height=0.35\linewidth]{figures/hoiem/rnn_layer2_x3rcnn_conv345normconcat_thresh0_sched2_in_0712tv/plots_diningtable_strong.pdf}
   \includegraphics[height=0.35\linewidth]{figures/hoiem/rnn_layer2_x3rcnn_conv345normconcat_thresh0_sched2_bbreg2wvall_bbthresh01abs_in_0712tv/plots_diningtable_strong.pdf}
   \includegraphics[height=0.35\linewidth]{figures/hoiem/rnn_layer2_x3rcnn_conv345normconcat_thresh0_nodrop_sched2_bbreg2wvall_bbthresh01abs_in_0712tv/plots_diningtable_strong.pdf}
  \end{center}
  \vspace{-12pt}
  \caption{%
    \textbf{VOC 2007 normalized AP for ``dining table.''}
  }
  \label{fig:hoiem}
\end{figure}

\begin{figure}[t]
  \begin{center}
   \includegraphics[height=0.35\linewidth]{figures/hoiem/noseg_conv5_thresh0_sched2_0712tv/plots_boat_strong.pdf}
   \includegraphics[height=0.35\linewidth]{figures/hoiem/noseg_conv5normconcat_thresh0_sched2_0712tv/plots_boat_strong.pdf}
   \includegraphics[height=0.35\linewidth]{figures/hoiem/noseg_conv45normconcat_thresh0_sched2_0712tv/plots_boat_strong.pdf}
   \includegraphics[height=0.35\linewidth]{figures/hoiem/noseg_conv345normconcat_thresh0_sched2_0712tv/plots_boat_strong.pdf}
   \includegraphics[height=0.35\linewidth]{figures/hoiem/rnn_layer2_x3rcnn_conv345normconcat_thresh0_sched2_in_0712tv/plots_boat_strong.pdf}
   \includegraphics[height=0.35\linewidth]{figures/hoiem/rnn_layer2_x3rcnn_conv345normconcat_thresh0_sched2_bbreg2wvall_bbthresh01abs_in_0712tv/plots_boat_strong.pdf}
   \includegraphics[height=0.35\linewidth]{figures/hoiem/rnn_layer2_x3rcnn_conv345normconcat_thresh0_nodrop_sched2_bbreg2wvall_bbthresh01abs_in_0712tv/plots_boat_strong.pdf}
  \end{center}
  \vspace{-12pt}
  \caption{%
    \textbf{VOC 2007 normalized AP for ``boat.''}
  }
  \label{fig:hoiem}
\end{figure}

\begin{figure}[t]
  \begin{center}
   \includegraphics[height=0.35\linewidth]{figures/hoiem/noseg_conv5_thresh0_sched2_0712tv/plots_bird_strong.pdf}
   \includegraphics[height=0.35\linewidth]{figures/hoiem/noseg_conv5normconcat_thresh0_sched2_0712tv/plots_bird_strong.pdf}
   \includegraphics[height=0.35\linewidth]{figures/hoiem/noseg_conv45normconcat_thresh0_sched2_0712tv/plots_bird_strong.pdf}
   \includegraphics[height=0.35\linewidth]{figures/hoiem/noseg_conv345normconcat_thresh0_sched2_0712tv/plots_bird_strong.pdf}
   \includegraphics[height=0.35\linewidth]{figures/hoiem/rnn_layer2_x3rcnn_conv345normconcat_thresh0_sched2_in_0712tv/plots_bird_strong.pdf}
   \includegraphics[height=0.35\linewidth]{figures/hoiem/rnn_layer2_x3rcnn_conv345normconcat_thresh0_sched2_bbreg2wvall_bbthresh01abs_in_0712tv/plots_bird_strong.pdf}
   \includegraphics[height=0.35\linewidth]{figures/hoiem/rnn_layer2_x3rcnn_conv345normconcat_thresh0_nodrop_sched2_bbreg2wvall_bbthresh01abs_in_0712tv/plots_bird_strong.pdf}
  \end{center}
  \vspace{-12pt}
  \caption{%
    \textbf{VOC 2007 normalized AP for ``bird.''}
  }
  \label{fig:hoiem}
\end{figure}